\documentclass{article} 
\usepackage{iclr2025_conference,times}

\usepackage{amsmath,amsfonts,bm}

\def\eqref#1{equation~\ref{#1}}

\def\1{\bm{1}}

\DeclareMathAlphabet{\mathsfit}{\encodingdefault}{\sfdefault}{m}{sl}
\SetMathAlphabet{\mathsfit}{bold}{\encodingdefault}{\sfdefault}{bx}{n}

\usepackage{hyperref}
\usepackage{url}
\usepackage{tabularx} 
\usepackage{booktabs}
\usepackage{multirow}
\usepackage{makecell}
\usepackage{enumitem}
\usepackage{algorithm}
\usepackage[noend]{algpseudocode}
\usepackage{graphicx}
\usepackage{wrapfig}
\usepackage{lipsum}  
\newcolumntype{x}[1]{>{\centering\arraybackslash}p{#1}}
\newcolumntype{Y}{>{\small\centering\arraybackslash}X}

\title{G2T-LLM: Graph-to-Tree Text Encoding for Molecule Generation with Fine-Tuned Large Language Models}

\author{Zhaoning Yu \\
    Iowa State University \\
    Ames, IA 50010 \\
    \texttt{znyu@iastate.edu}
    \And Xiangyang Xu \\
    Iowa State University \\
    Ames, IA 50010 \\
    \texttt{xyxu@iastate.edu}
    \And Hongyang Gao \\
    Iowa State University \\
    Ames, IA 50010 \\
    \texttt{hygao@iastate.edu}
}

\iclrfinalcopy 
\begin{document}

\maketitle

\begin{abstract}
We introduce G2T-LLM, a novel approach for molecule generation that uses graph-to-tree text encoding to transform graph-based molecular structures into a hierarchical text format optimized for large language models (LLMs). This encoding converts complex molecular graphs into tree-structured formats, such as JSON and XML, which LLMs are particularly adept at processing due to their extensive pre-training on these types of data. By leveraging the flexibility of LLMs, our approach allows for intuitive interaction using natural language prompts, providing a more accessible interface for molecular design. Through supervised fine-tuning, G2T-LLM generates valid and coherent chemical structures, addressing common challenges like invalid outputs seen in traditional graph-based methods. While LLMs are computationally intensive, they offer superior generalization and adaptability, enabling the generation of diverse molecular structures with minimal task-specific customization. The proposed approach achieved comparable performances with state-of-the-art methods on various benchmark molecular generation datasets, demonstrating its potential as a flexible and innovative tool for AI-driven molecular design.
\end{abstract}

\section{Introduction}

Molecular generation is a critical task in fields such as drug discovery, material science, and chemistry \citep{111&121, 112, 113}. The ability to design and create novel molecules with specific properties can accelerate the development of new therapies, advanced materials, and innovative chemicals. Traditional approaches to molecular generation, such as rule-based systems \citep{111&121, 122} and graph-based \citep{you2018graphrnn, madhawa2019graphnvp, shi2020graphaf} models, have provided foundational tools. However, these methods often face limitations in generating diverse, valid, and chemically coherent molecular structures, restricting their ability to explore the vast chemical space effectively \citep{vignac2022digress, jo2022score}.
Recent advancements in deep learning, especially the rise of large language models (LLMs), offer new opportunities for molecular generation \citep{ brahmavar2024generating, wang2024grammar, yao2024exploring}. Unlike traditional methods, LLMs are not constrained by domain-specific rules and can generalize from vast amounts of data. This flexibility allows them to generate creative and diverse content, potentially uncovering novel chemical compounds. Prior non-LLM approaches, such as graph-based generative models \citep{you2018graphrnn, madhawa2019graphnvp, shi2020graphaf, luo2021graphdf, vignac2022digress, jo2022score}, often struggle with limited generalization, rule-based rigidity, or difficulty scaling to more complex chemical structures. In contrast, LLMs can adapt to a wide range of prompts and provide greater flexibility, making them an attractive choice for AI-driven molecular generation.

Despite the promise of LLMs, applying them to molecular generation presents a unique challenge. Molecular structures are typically represented as graphs, with atoms as nodes and bonds as edges. LLMs, however, are trained to understand sequences of tokens \citep{vaswani2017attention}, particularly in structured text formats such as XML and JSON \citep{brown2020language}, and are not inherently designed to process graph-based data. This mismatch creates a barrier when attempting to use LLMs for tasks that require understanding the relational and non-linear properties of molecular structures. LLMs \citep{luo2023molfm, le2024molx} may struggle to generate chemically valid or meaningful molecules without proper representation.

To overcome this challenge, we propose a novel Graph-to-Tree Text Encoding designed to transform molecular graphs into a format that LLMs can process effectively. Inspired by SMILES but not relying on it, our encoding converts graph-based molecular structures into hierarchical text representations, such as JSON and XML. These formats are naturally suited to LLMs, which excel at interpreting tree-like structures due to their training on similar data. By converting molecular graphs into tree-structured text, we align the data representation with the strengths of LLMs, enabling them to understand and generate molecules more effectively.
With the graph-to-tree text encoding in place, we supervised fine-tuned LLMs to generate valid and coherent chemical structures. This fine-tuning process ensures that the generated molecules adhere to chemical rules and constraints, addressing common challenges such as the generation of invalid or chemically infeasible molecules. The fine-tuning allows LLMs to learn how to translate natural language prompts into meaningful molecular designs, opening new possibilities for human-guided molecule generation.
Our approach has demonstrated comparable performances with state-of-the-art (SOTA) models on several benchmark molecular generation datasets. These results validate the effectiveness of our graph-to-tree encoding in making LLMs capable of generating chemically sound and diverse molecules. Additionally, the performance gains achieved underscore the potential of LLMs as a flexible and innovative tool for molecular generation, particularly when paired with a well-suited encoding.

This work makes the following contributions:
\begin{itemize}[leftmargin=10pt]
    \item We propose G2T-LLM, a novel approach that transforms graph-based molecular structures into text formats like JSON and XML, optimized for large language models.
    \item We introduce a token constraining technique to guide the LLM’s generation process, ensuring that the output adheres to the expected tree-structured format, which is critical for maintaining molecular coherence.
    \item We develop a supervised fine-tuning method to enable LLMs to generate valid and coherent chemical structures, leveraging graph-to-tree text encoding.
    \item We achieve comparable performances with state-of-the-art models on benchmark molecular generation datasets, demonstrating the effectiveness and potential of our approach for AI-driven molecular design.
\end{itemize}

\section{Related Work}
\textbf{Graph Generation.}
The graph generation task aims to learn the distribution of graphs. The traditional approaches \citep{zang2020moflow, shi2020graphaf, luo2021graphdf, you2018graphrnn, madhawa2019graphnvp, dai2018syntax} such as auto-regression, Generative Adversarial Network (GAN), and Variational Autoencoder (VAE) have been explored for this purpose. However, they have faced challenges in modeling the permutation-invariant nature of graph distribution and learning the relationship between edges and nodes, often due to limitations in their model capacity.
Recent advancements in diffusion methods \citep{niu2020permutation,jo2022score,vignac2022digress, jo2023graph} have significantly improved graph generation. GDSS \citep{jo2022score} generates both node features and adjacency matrices simultaneously, resulting in better alignment with graph datasets. DiGress \citep{vignac2022digress} addresses the challenge of generating graphs with categorical node and edge attributes, which is a difficult task due to the unordered nature and sparsity of graphs. GruM \citep{jo2023graph} directly learns graph topology, improving connectivity and structure recovery.

\textbf{Graph to Text for LLM.}
The emergence of large language models (LLMs) has driven significant advancements in the natural sciences \citep{taylor2022galactica, liu2024moleculargpt}. These models are trained on vast amounts of text data, the most abundant type of data, contributing to their success across many tasks. 
Multi-modal methods \citep{luo2023molfm, le2024molx} have been proposed to incorporate both graph and text information. They typically rely on graph neural networks or transformers to encode graphs. However, these methods often use text, such as SMILES, to represent molecular features. SMILES may not tokenize the molecular structure effectively, limiting the ability to represent the molecule structure accurately. As a result, the graph embeddings may be too weak for intricate molecular structures, limiting performance in molecular generation tasks.

Recently, there have been attempts \citep{fatemi2023talk} to represent graphs in natural language formats, encoding their structure using descriptive language. However, this naive approach introduces challenges, as such encodings are unlikely to appear in typical text, meaning that LLMs—trained predominantly on conventional text data—may struggle to process them effectively. Using an encoding that aligns with the LLMs' training data is essential. We propose leveraging tree-structured formats like JSON and XML to encode molecules to address this issue. The JSON format is a widely used and structured data representation commonly found in LLM training. This allows us to capture the complexity of molecular graphs while ensuring compatibility with LLMs.

\section{G2T-LLM}

This section introduces G2T-LLM: Graph-to-Tree Text Encoding for Molecule Generation with Fine-Tuned Large Language Models.

\begin{figure*}[t]
\center
\label{fig:encode}
\includegraphics[width=\textwidth]{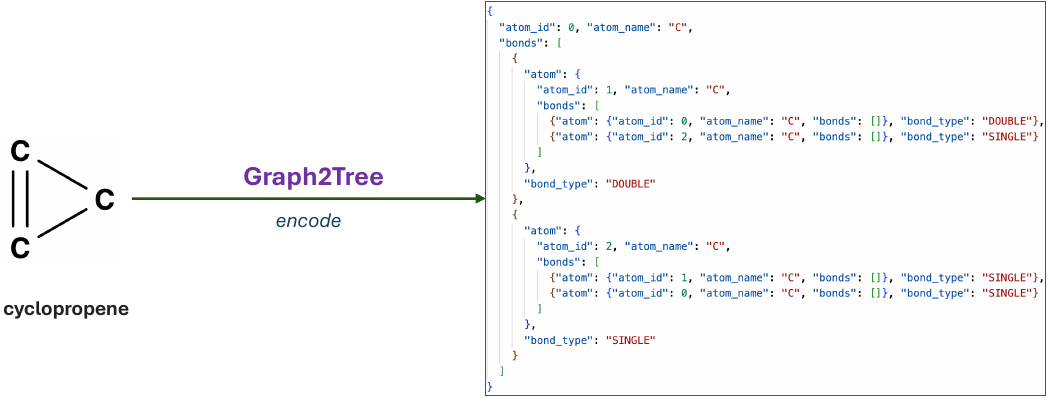}
\caption{Illustration of the Graph-to-Tree Text Encoding process described in Section~\ref{sec:graph2tree} and Algorithm~\ref{alg:graph2tree}. This figure shows how the molecular structure of cyclopropene is transformed into a hierarchical tree representation. Each atom and bond is mapped to nodes and edges in the tree, with unique identifiers assigned.
}
\end{figure*}

\subsection{Challenges and Motivations}

Molecular graphs pose a challenge for LLMs due to their inherently complex, non-linear structures, where atoms (nodes) and bonds (edges) form intricate connectivity patterns, including rings, branches, and cycles. Traditional LLMs excel at processing sequential data, such as natural language, where information flows in a linear manner. However, molecular graphs do not naturally conform to this format, as their connections often lack a clear, ordered sequence. This mismatch complicates the application of LLMs to molecule-related tasks.

Despite these challenges, LLMs have shown a capacity to handle structured, hierarchical data formats, such as JSON and XML. These formats share some of the complexity of graphs but are still expressed as trees, with clear parent-child relationships between elements. LLMs trained on such data can handle hierarchical structures by processing them as sequences while maintaining the relationships and nested dependencies inherent to these structures. This training has made LLMs particularly adept at handling data that can be decomposed into nested layers, making them better suited for tree-like representations than arbitrary graphs.

To leverage this strength, we propose encoding molecular graphs into a tree structure. This approach is inspired by SMILEs, which are essentially tree representations of molecular graphs, proving that molecular graphs can be effectively serialized as trees while preserving their chemical properties. This encoding acts as a bridge between the graph-based molecular structures and the LLM’s ability to process and generate hierarchical data. The LLM can be trained on these tree-encoded molecules, and it can also output molecules in the same structured format, facilitating the generation of coherent molecular representations.
By aligning graph data with a format that LLMs are well-equipped to handle, this method holds the potential for improving the coherence and plausibility of generated molecular structures.

\subsection{Graph-to-Tree Text Encoding}\label{sec:graph2tree}

To make molecular graphs accessible to LLMs, we introduce a tree-based encoding inspired by the SMILES format. SMILES encodes molecules by performing a depth-first traversal over the molecular graph and representing it as a linear string. In our approach, we extend this traversal to build a hierarchical tree structure, where atoms are represented as nodes and their bonds as edges connecting them. The hierarchical nature of the tree is well-suited for the LLM’s training with tree-like structures.

However, molecular graphs often contain rings and cycles—features that trees cannot naturally represent. To address this, we assign each atom in the molecule a unique identifier (ID). When the traversal encounters a ring closure or cycle, the tree refers back to the atom’s unique ID rather than creating a new node, thereby preserving both the hierarchical structure and chemical validity. This encoding technique ensures that we accurately capture the full molecular graph in a way the LLM can process, while maintaining the integrity of complex molecular features such as rings and branches. Algorithm~\ref{alg:graph2tree} and Algorithm~\ref{alg:tree2graph} describe the processes for converting a molecular graph to a tree-structured text representation and for reconstructing the graph from this format, respectively. Figure~\ref{fig:encode} illustrates the graph-to-tree text encoding.

\begin{algorithm}[t]
\caption{Convert Molecular Graph to Tree-Structured Text Representation}
\label{alg:graph2tree}
\begin{algorithmic}[1]
    \Function{Graph2Tree}{graph}
        \State \textbf{Input:} graph (dictionary of atom identifiers to connected atom identifiers)
        \State \textbf{Output:} text\_representation (tree structure in text format)
        \State tree $\gets$ \{\} \Comment{Initialize tree}
        \State visited $\gets$ \{\} \Comment{Set to track visited atoms}
        \State unique\_id\_counter $\gets 0$ \Comment{Counter for unique atom IDs}
        \State id\_mapping $\gets$ \{\} \Comment{Mapping of atoms to unique IDs}

        \Function{ConvertAtom}{atom}
            \State visited.add(atom)
            \State atom\_id $\gets$ unique\_id\_counter
            \State id\_mapping[atom] $\gets$ atom\_id
            \State unique\_id\_counter $\gets$ unique\_id\_counter + 1
            \State bonds $\gets$ []
            \For{neighbor, bond\_type \textbf{in} graph[atom]}
                \If{neighbor $\notin$ visited}
                    \State child $\gets$ \Call{ConvertAtom}{neighbor}
                \Else
                    \State neighbor\_id $\gets$ id\_mapping[neighbor]
                    \State child $\gets$ \{``atom\_name'': atom.atom\_name, ``atom\_id'': neighbor\_id, ``bonds'': []\} 
                    \State\Comment{Set bonds to empty to avoid circular references}
                \EndIf
                \State bonds.append(\{``atom'': child. ``bond\_type'': bond\_type\})
            \EndFor
            \State \Return \{``atom\_name'': atom.atom\_name, ``atom\_id'': atom\_id, ``bonds'': bonds\}
        \EndFunction

        \State root\_atom $\gets$ any(graph.keys()) \Comment{Start from any atom as the root}
        \State tree $\gets$ \Call{ConvertAtom}{root\_atom}

        \State text\_representation $\gets$ JSON.stringify(tree) \Comment{Convert tree to JSON text format}
        \State \Return text\_representation
    \EndFunction
\end{algorithmic}
\end{algorithm}

\begin{algorithm}[t]
\caption{Convert Tree-Structured Text to Molecular Graph}
\label{alg:tree2graph}
\begin{algorithmic}[1]
    \Function{Tree2Graph}{tree\_json}
        \State \textbf{Input:} tree\_json (tree structure in JSON format)
        \State \textbf{Output:} graph (dictionary representing the molecular graph)
        \State tree $\gets$ JSON.parse(tree\_json) \Comment{Convert JSON text to tree structure}
        \State graph $\gets$ \{\} \Comment{Initialize graph structure}

        \Function{ConvertNodeToGraph}{node, parent, bond\_type}
            \State atom\_id $\gets$ node[``atom\_id'']
            \If{atom\_id $\in$ id\_mapping}
                \State atom $\gets$ id\_mapping[atom\_id]
            \Else
                \State atom\_name $\gets$ node[``atom\_name'']
                \State atom $\gets$ new Node(atom\_name)
                \State id\_mapping[atom\_id] $\gets$ atom
                \State graph[atom] $\gets$ [] \Comment{Initialize adjacency list}
            \EndIf

            \If{parent\_id $\neq$ null}
                \State graph[atom].append((parent, bond\_type))
                \State graph[parent].append((atom, bond\_type))
            \EndIf

            \For{child \textbf{in} node[``bond'']}
                \State \Call{ConvertNodeToGraph}{child, atom}
            \EndFor
        \EndFunction

        \State root\_node $\gets$ tree \Comment{Start with the root node of the tree}
        \State \Call{ConvertNodeToGraph}{root\_node, null}
        
        \State \Return graph
    \EndFunction
\end{algorithmic}
\end{algorithm}

\subsection{Token Constraining for Valid Tree-Structure Generation}\label{sec:3.3}

Despite the advancements in LLMs, there remains a significant challenge in ensuring that the outputs adhere to valid tree-structured formats. LLMs, while capable of generating coherent text, may produce sequences that do not respect the hierarchical relationships required for molecular representation. This can lead to outputs that are structurally invalid, failing to accurately represent the complex relationships inherent in molecular graphs.

To mitigate this issue, we implement a set of constraints that guide the token generation process of the LLM. These constraints filter the tokens allowed at each step, ensuring that generated outputs remain within the bounds of valid tree structures. Specifically, we impose rules that dictate acceptable parent-child relationships, enforce valid connections between atoms, and restrict the formation of non-hierarchical sequences. Additionally, we constrain the types of atoms and bonds that can be generated, ensuring that only valid atom types (e.g., carbon, oxygen) and bond types (e.g., single, double) are used in the output. This approach leverages domain knowledge of molecular structures to create a robust framework for guiding the LLM's outputs.

The application of token constraining significantly enhances the reliability of the generated tree-structured outputs. By enforcing these constraints, we improve the chances that the LLM produces valid representations of molecular structures that can be effectively used in further analyses or applications. This technique not only aids in ensuring the accuracy of the generated data but also reinforces the overall effectiveness of our graph-to-tree text encoding approach, making it a vital component in achieving coherent and chemically valid molecular generation.

\subsection{Supervised Fine-Tuning LLMs for Molecular Generation}

A key challenge in leveraging large language models for molecular generation is that, without specialized training, they may struggle to produce valid molecular structures, particularly when dealing with complex features such as rings, cycles, and the inherent chemical constraints that govern molecular formation. Supervised fine-tuning addresses this issue by teaching the LLM domain-specific rules and patterns, enabling it to generate valid molecular structures that adhere to chemical principles.

\begin{figure*}[t]
\center
\label{fig:finetune}
\includegraphics[width=\textwidth]{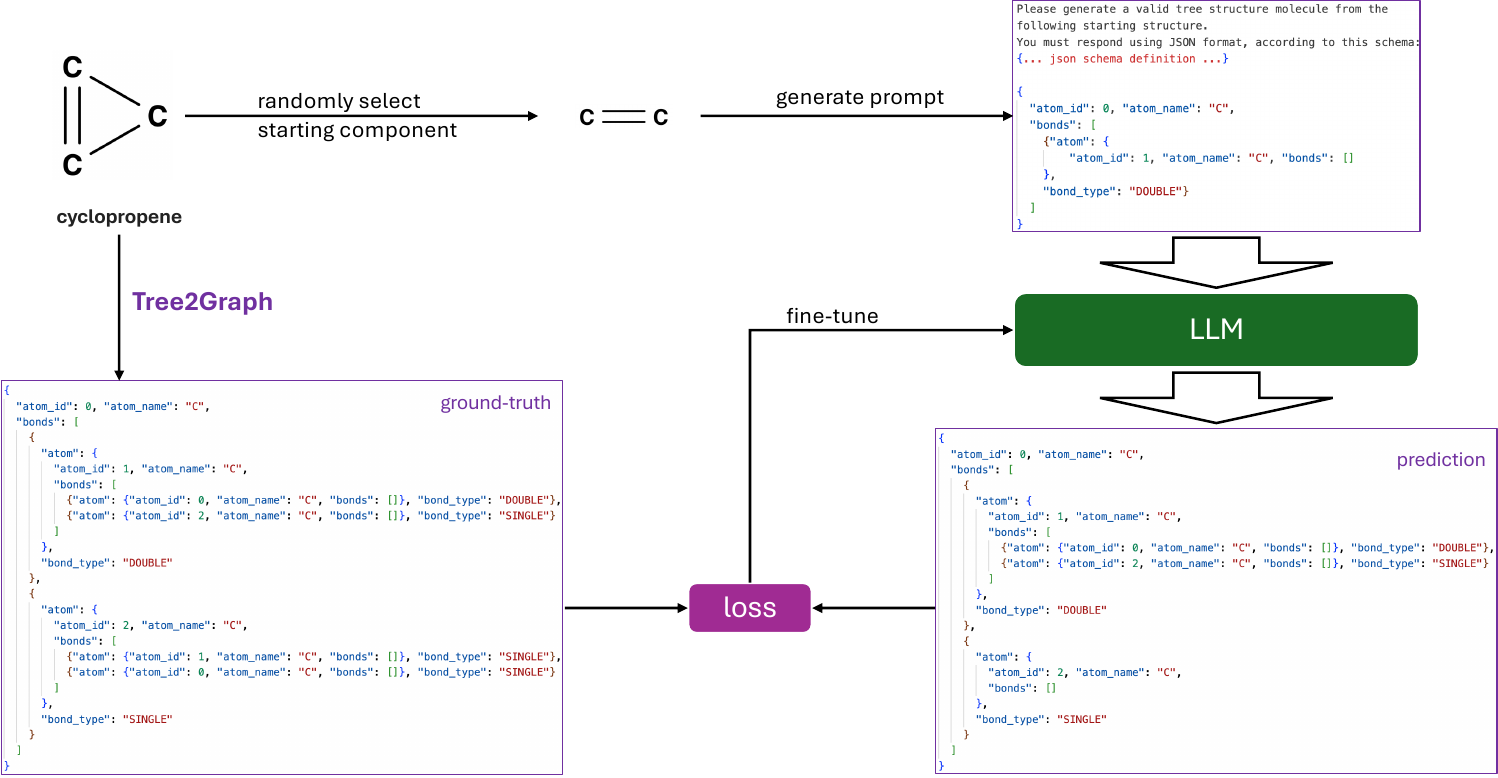}
\caption{An illustration of the supervised fine-tuning process of G2T-LLM. The process begins by randomly selecting a starting component, exemplified by cyclopropene, which is encoded into a partial tree structure and passed as a prompt to the LLM. The LLM generates the remaining molecular structure, which is compared against the ground truth. A loss is computed and is used to fine-tune the model, iteratively improving its performance in generating valid molecular graphs.}
\end{figure*}

We structure the fine-tuning process as a molecular completion task. The LLM is trained by prompting it with a partial molecular structure, encoded using the graph-to-tree text encoding and tasking it with predicting the remaining atoms and bonds necessary to complete the molecule. For each training example, we provide the LLM with an incomplete molecular graph, and the model is then expected to generate the missing parts based on the information provided. The model's output is evaluated against the full molecular structure’s text encoding, and the loss is computed based on the accuracy of its predictions. By iterating through this process, the LLM learns to predict the completion of molecular graphs in a way that respects chemical validity, helping the model better handle challenging structural features. Note that token constraining is deliberately omitted during fine-tuning, allowing the LLM to explore and learn more freely before constraints are imposed during inference. Figure~\ref{fig:finetune} illustrates the supervised fine-tuning process of G2T-LLM.

The fine-tuning process is integral to the success of our approach. By casting molecular generation as a completion task and using the proposed graph-to-tree encoding as a bridge between molecular structures and the LLM’s capabilities, we enhance the model’s ability to generate coherent and chemically valid outputs. This fine-tuning approach refines the LLM's understanding of molecular patterns and constraints, enabling it to produce outputs that are more reliable and scientifically grounded within the realm of molecular design.

\begin{figure*}[t]
\center
\label{fig:inference}
\includegraphics[width=\textwidth]{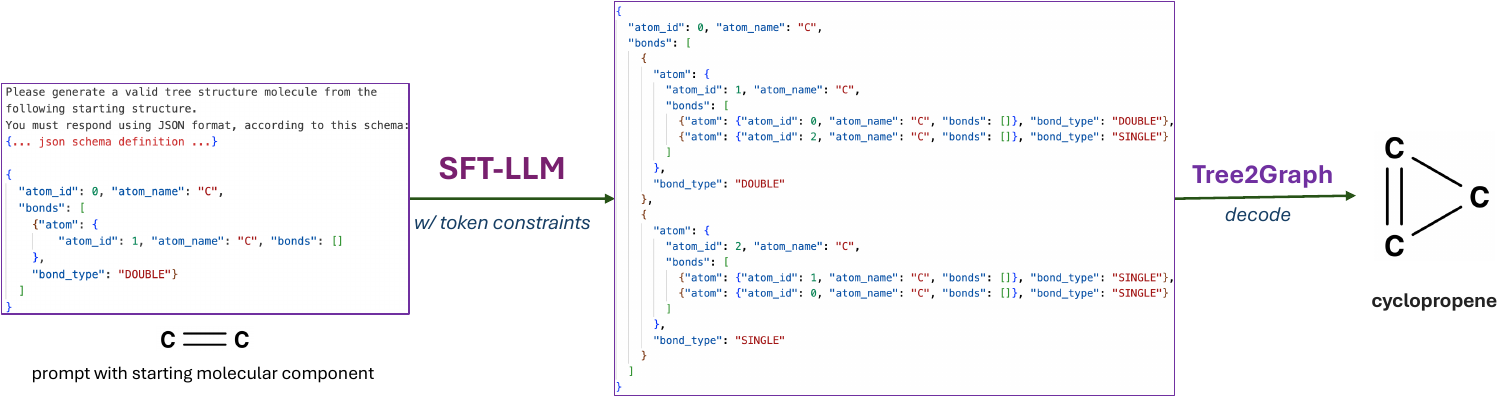}
\caption{An illustration of the inference process of G2T-LLM. The process starts by prompting the model with a random molecular component. The model, a fine-tuned LLM (SFT-LLM), generates new molecular structures while applying token constraints to ensure valid outputs. The output is a tree-structured text representing the molecule. It is then decoded back into a molecular graph corresponding to cyclopropene.}
\end{figure*}

\subsection{Inference Process of G2T-LLM}

The molecular generation process begins with selecting a random molecular component, which could be an atom, a bond, or even a larger motif. This component serves as the initial prompt for the fine-tuned LLM. The component is encoded into the graph-to-tree text format, creating a tree-structured representation that the LLM can process.

Once the LLM receives this initial prompt, it is tasked with generating the subsequent components of the molecular structure. At each step, the LLM's output is constrained by the Token Constraining mechanism, ensuring that only chemical and schema-valid tokens—such as specific atom types and bond types—are generated. These constraints help guide the LLM in maintaining the coherence of the structure, preventing invalid or nonsensical outputs, and ensuring that the generated molecule adheres to the expected chemical rules.
As the LLM iteratively predicts new components, these outputs are progressively combined into an expanding tree-structured text. This generated text represents the molecular graph, with nodes corresponding to atoms and edges corresponding to bonds. Once the generation process is complete, the final tree-structured text is decoded back into a full molecular graph. This graph is then translated into a standard molecular format, fully reconstructing the molecule from the text generated by the LLM. Figure~\ref{fig:inference} illustrates the inference process of G2T-LLM.
\section{Experiments}

In this section, we conduct comprehensive experiments on two real-world datasets to evaluate the effectiveness of our proposed methods. 
\subsection{Experimental Setup}
\textbf{Datasets and Metrics.} We evaluate the quality of molecule generation using two real-world datasets: \textit{QM9}~\citep{ramakrishnan2014quantum} and \textit{ZINC250k}~\citep{irwin2012zinc}. Following the evaluation setting used in~\citep{jo2023graph}, we measure model performance across four metrics. 
\textit{Validity} is the proportion of generated molecules that are valid without any valency corrections.
\textit{Novelty} is the proportion of valid molecules that are not present in the training dataset.
\textit{Frechet ChemNet Distance (FCD)}~\citep{preuer2018frechet} measures the similarity between two molecule sets by comparing the activations of the penultimate layer of the ChemNet model. 
\textit{Scaffold similarity (Scaf.)} evaluates the model's ability to generate similar substructures.

\textbf{Baselines.} We compare our model with following molecular graph generation methods. \textit{MoFlow}~\citep{zang2020moflow} is a one-shot flow-based model that generates entire molecular graphs in a single step. \textit{GraphAF}~\citep{shi2020graphaf} and \textit{GraphDF}\citep{luo2021graphdf} are autoregressive flow-based models, generating molecules sequentially. Additionally, we evaluate against the diffusion models. \textit{EDP-GNN}~\citep{niu2020permutation} is a score-based model designed for generating adjacency matrices. \textit{GDSS}~\citep{jo2022score} uses a continuous diffusion process for molecule generation, \textit{DiGress}~\citep{vignac2022digress} employs a discrete diffusion approach, and \textit{Grum}~\citep{jo2023graph} designed a mixture of endpoint-conditioned diffusion processes.


Although several studies have explored using LLMs for molecular generation, direct comparisons with our approach are not feasible. For instance, LMLF \citep{brahmavar2024generating}, Grammar Prompting \citep{wang2024grammar}, and LLM4GraphGen \citep{yao2024exploring} all employ rule-based prompt-engineering techniques that fundamentally differ from our SFT LLM approach. These models rely on predefined rules and heuristics to guide the generation process, which restricts their ability to learn from the underlying data distributions. In contrast, our method leverages a more flexible and adaptive encoding, allowing the LLM to capture the complexities of molecular structures more effectively.

Moreover, the baseline models utilize significantly larger architectures, such as GPT-4, whereas our experiments are conducted with LLaMA3.1-8B. This disparity in model size and complexity further complicates direct comparisons, as the performance capabilities and learned representations of these models can vary widely. Therefore, assessing our results against those achieved by larger, rule-based models may not provide a meaningful evaluation of performance, given the substantial differences in methodologies and model architectures.

\textbf{Implementation details.}
For our G2T-LLM, we conduct experiments using the LLaMA3.1-8B model \citep{dubey2024llama} as our base LLM, selected for its strong performance in text generation tasks. The model parameters are fine-tuned with torchtune \citep{Ansel_PyTorch_2_Faster_2024}, and we leverage QLoRA \citep{dettmers2024qlora} to accelerate training while reducing memory consumption. The fine-tuning dataset consists of 5,000 molecules, and the model is trained with a batch size of 8, using the AdamW optimizer \citep{loshchilov2017decoupled} with a weight decay of 0.01 and a learning rate of 3e-4. The learning rate is adjusted by a cosine schedule with 100 warmup steps, and cross-entropy loss is employed for the loss computation. All model computations are performed with the bf16 data type. Fine-tuning is carried out on an NVIDIA A100 SXM4 80GB, and inference is done on NVIDIA GeForce RTX 3090 and 4090 GPUs. The implementation is done in PyTorch~\citep{paszke2019pytorch}.

\subsection{Experimental Results}

\begin{table}[t]
    \centering
    \caption{Generation results on the QM9 and ZINC250k datasets. We report the mean of 3 different runs. The best results are highlighted in bold. The second-best results are highlighted in \underline{underline}. We provide the results of uniqueness, and NSPDK in Appendix~\ref{app:res}.}
    \label{tab:exp}
    \begin{tabularx}{\linewidth}{l|YYYY|YYYY}
    \toprule
    Datasets  & \multicolumn{4}{c}{QM9} & \multicolumn{4}{c}{ZINC250K} \\ \midrule
    Methods  & Valid$\uparrow$ & Novelty$\uparrow$  & FCD$\downarrow$  & Scaf$\uparrow$ & Valid$\uparrow$ & Novelty$\uparrow$  & FCD$\downarrow$  & Scaf$\uparrow$ \\ \midrule

    MoFlow            & 91.36 & \underline{94.72} & 4.467 & 0.1447 & 63.11 & \textbf{100.00} & 20.931 & 0.0133 \\
    GraphAF           & 74.43 & 86.59 & 5.625 & 0.3046 & 68.47 & 99.99 & 16.023 & 0.0672 \\
    GraphDF           & 93.88 & \textbf{98.54} & 10.928 & 0.0978 & 90.61 & \textbf{100.00} & 33.546 & 0.0000 \\ \midrule
    EDP-GNN           & 47.52 & 86.58 & 2.680 & 0.3270 & 82.97 & \textbf{100.00} & 16.737 & 0.0000 \\
    GDSS              & 95.72 & 86.27 & 2.900 & 0.6983 & 97.01 & \textbf{100.00} & 14.656 & 0.0467 \\
    DiGress           & 98.19 & 25.58  & \textbf{0.095} & \underline{0.9353} & 94.99 & 99.99 & 3.482 & 0.4163 \\
    Grum              & \textbf{99.69} & 24.15 & \underline{0.108} & \textbf{0.9449} & \textbf{98.65} & 99.98 & \textbf{2.257} & \underline{0.5299} \\ \midrule
    Ours              & \underline{99.47} & 88.29 & 0.815 & 0.9112 & \underline{98.03} & \textbf{100.00} & \underline{2.445} & \textbf{0.6062} \\ \bottomrule
    \end{tabularx}
\end{table}

\begin{figure*}[t]
    \centering
    \scriptsize
    \begin{tabularx}{\textwidth}{c| c| c| c | c|c| c| c | c}
        \hline
        QM9 Ref. & Ours & Grum & GDSS & GDSS-seq 
        & GraphAF & MoFlow & GraphDF & EDP-GNN \\
        \hline
        \multicolumn{1}{c|}{\includegraphics[width=0.08\textwidth]{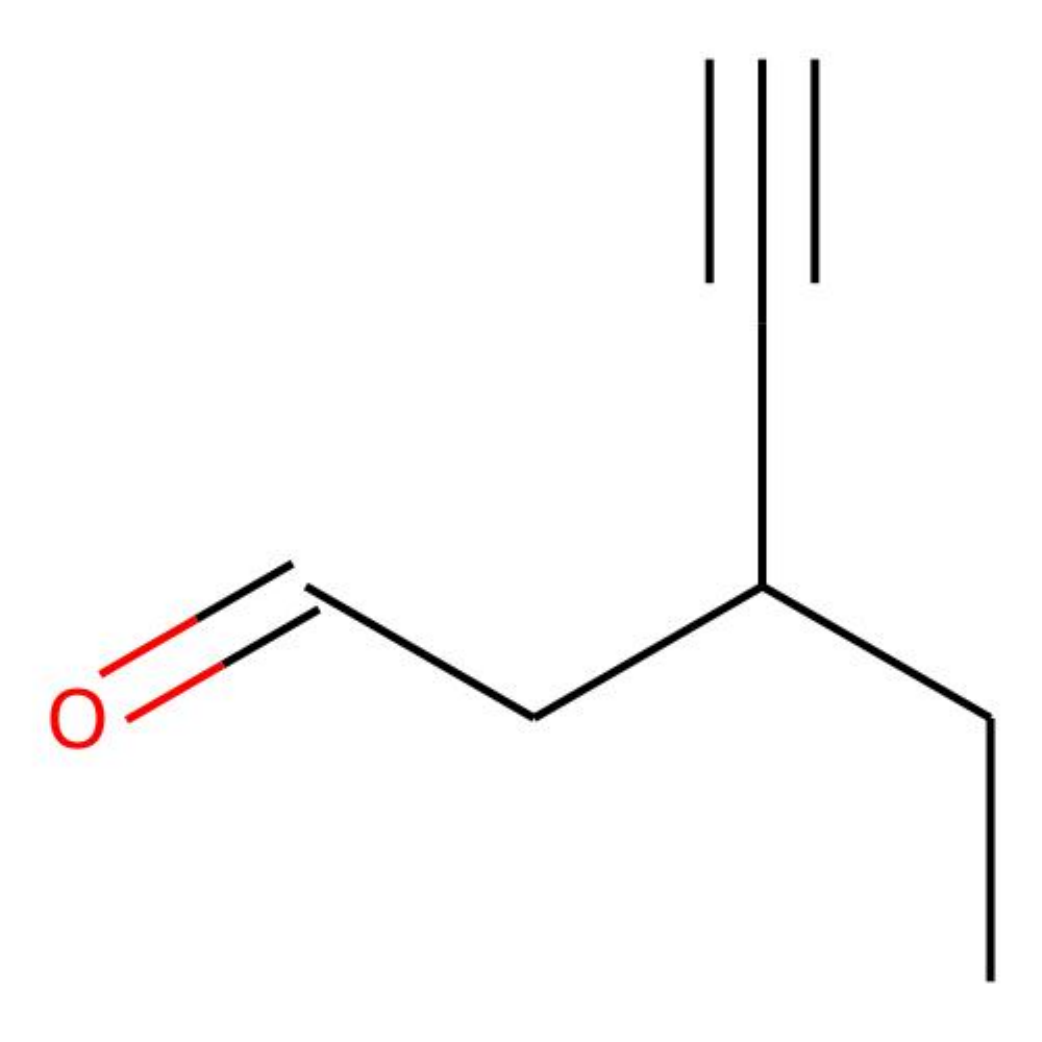}} &
        \multicolumn{1}{c|}{\includegraphics[width=0.08\textwidth]{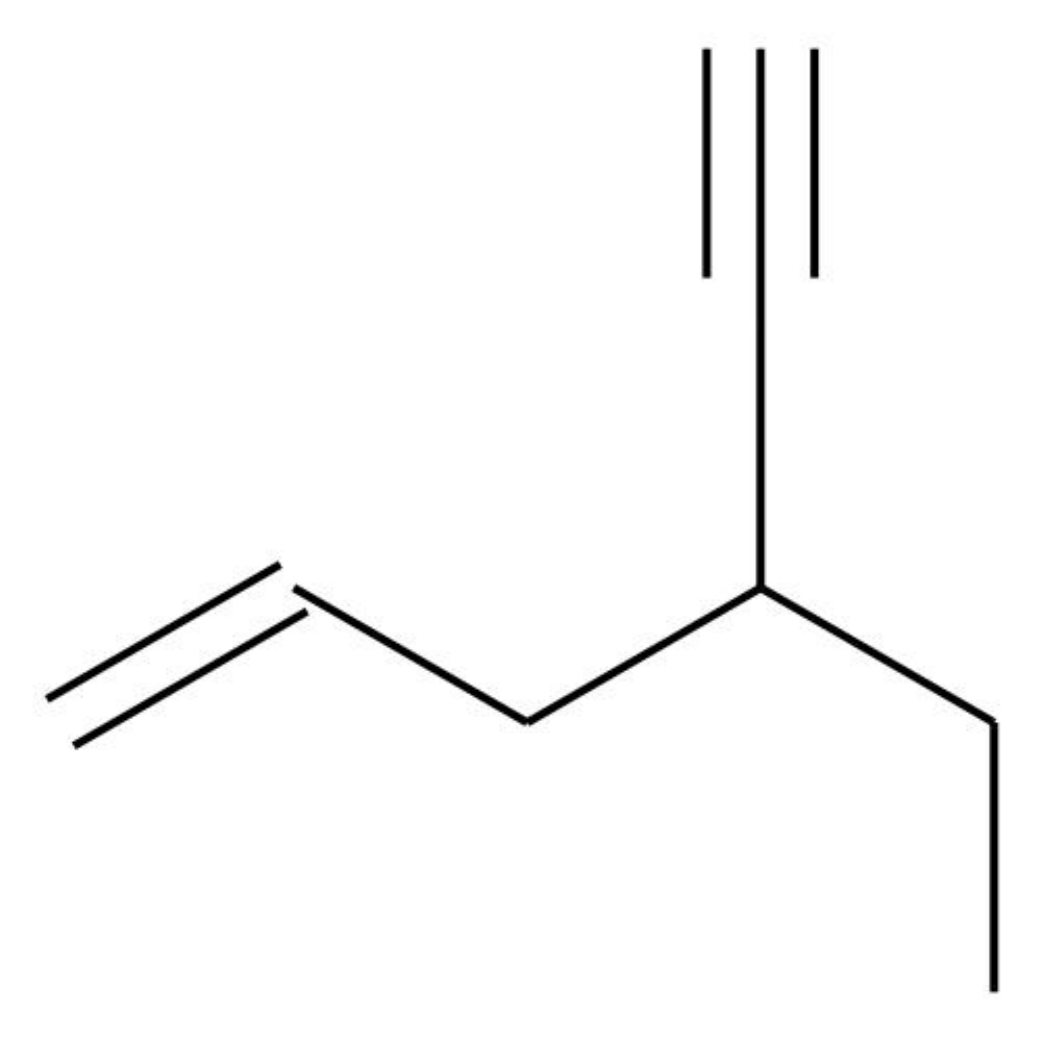}} &
        \multicolumn{1}{c|}{\includegraphics[width=0.08\textwidth]{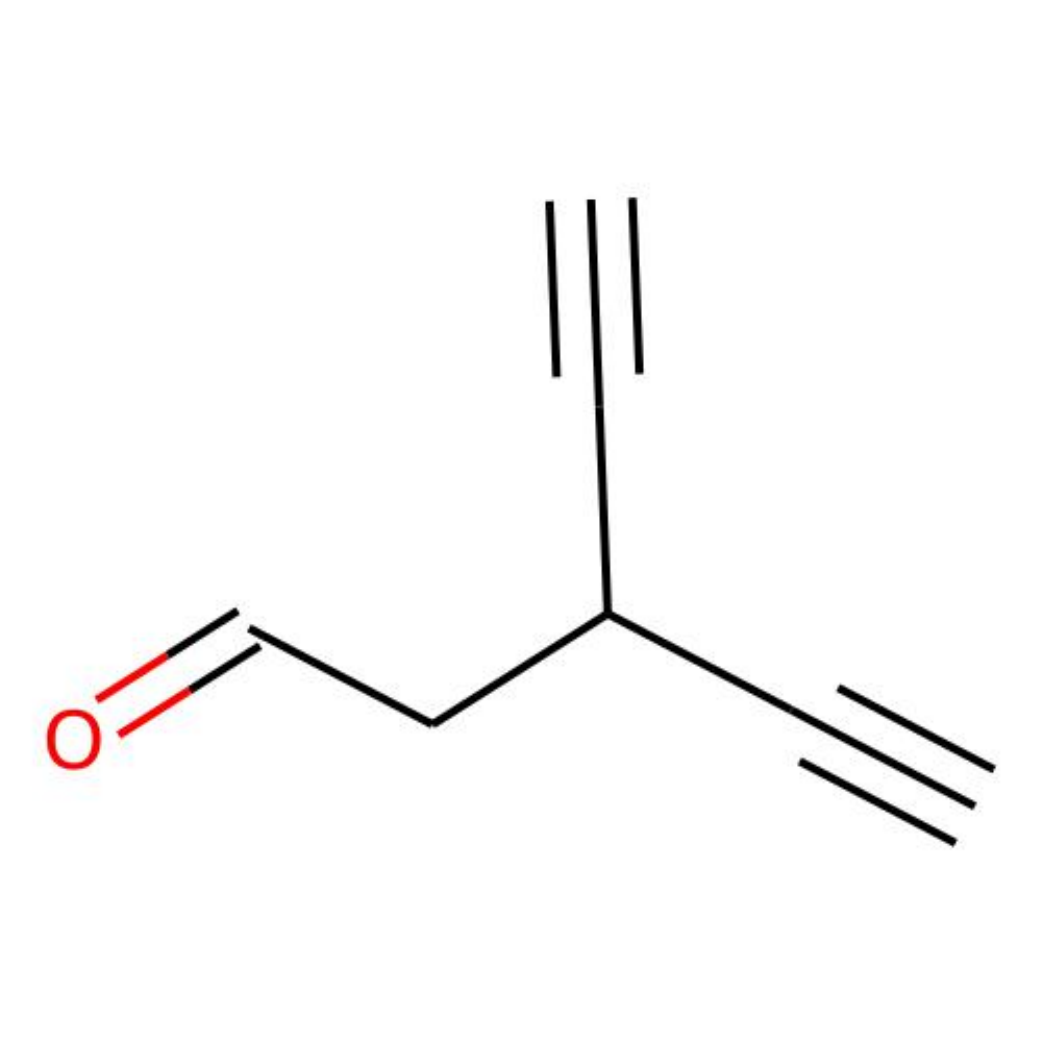}}&
        \multicolumn{1}{c|}{\includegraphics[width=0.08\textwidth]{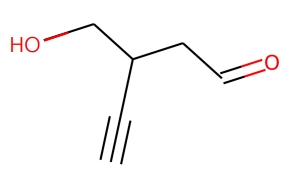}} &
        \multicolumn{1}{c|}{\includegraphics[width=0.08\textwidth]{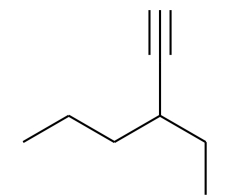}} &
        \multicolumn{1}{c|}{\includegraphics[width=0.08\textwidth]{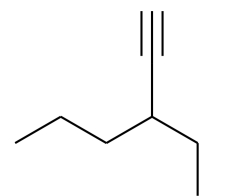}} &
        \multicolumn{1}{c|}{\includegraphics[width=0.08\textwidth]{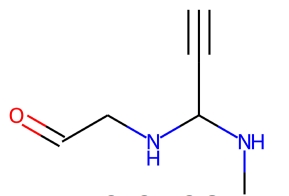}}&
        \multicolumn{1}{c|}{\includegraphics[width=0.08\textwidth]{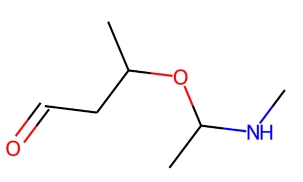}}&
        \multicolumn{1}{c}{\includegraphics[width=0.08\textwidth]{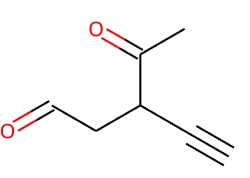}}\\
        similarity & \textbf{0.6000} & 0.5000 & \textbf{0.6000} & 
        0.4800 & 0.4800 & 0.3438 
        & 0.2727 & 0.4483 \\
        \hline
        
        \multicolumn{1}{c|}{\includegraphics[width=0.08\textwidth]{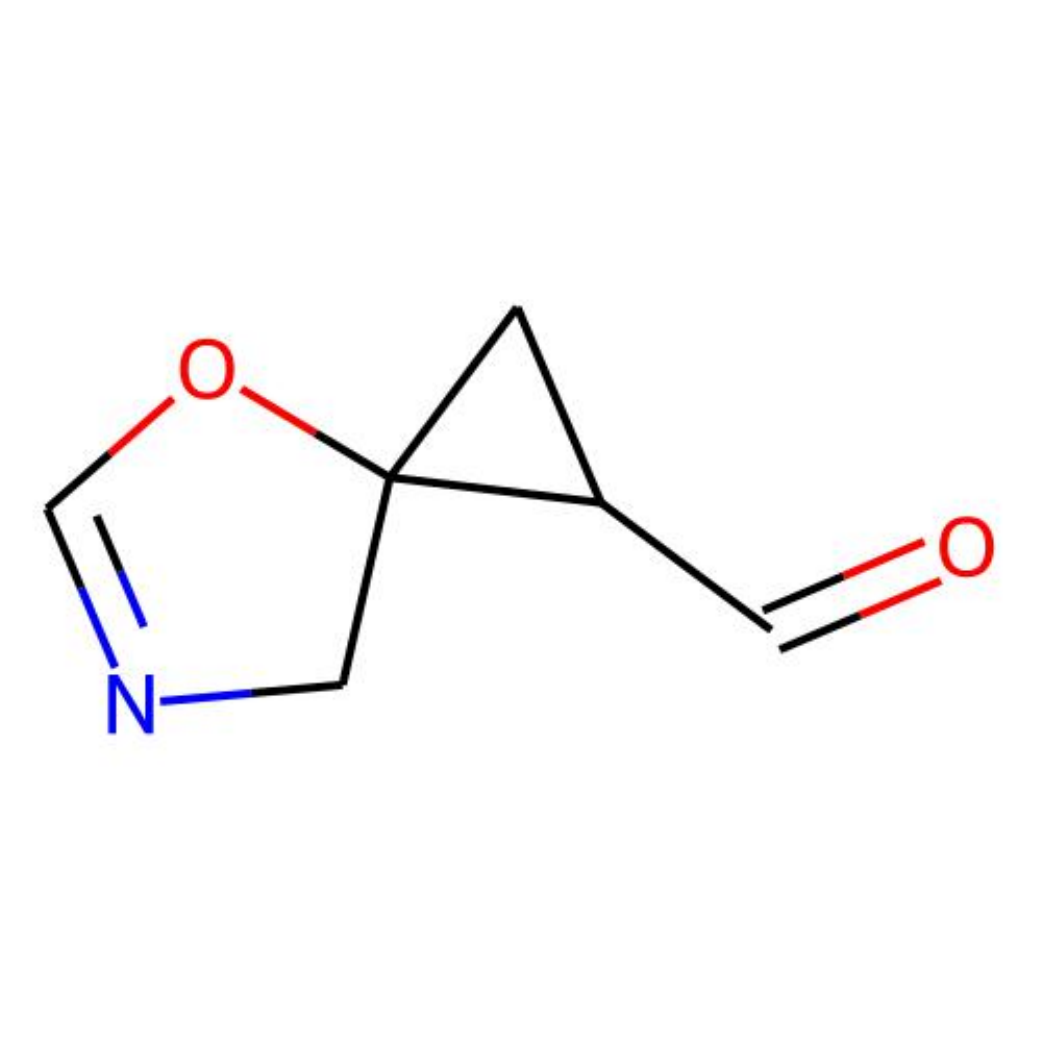}} &
        \multicolumn{1}{c|}{\includegraphics[width=0.08\textwidth]{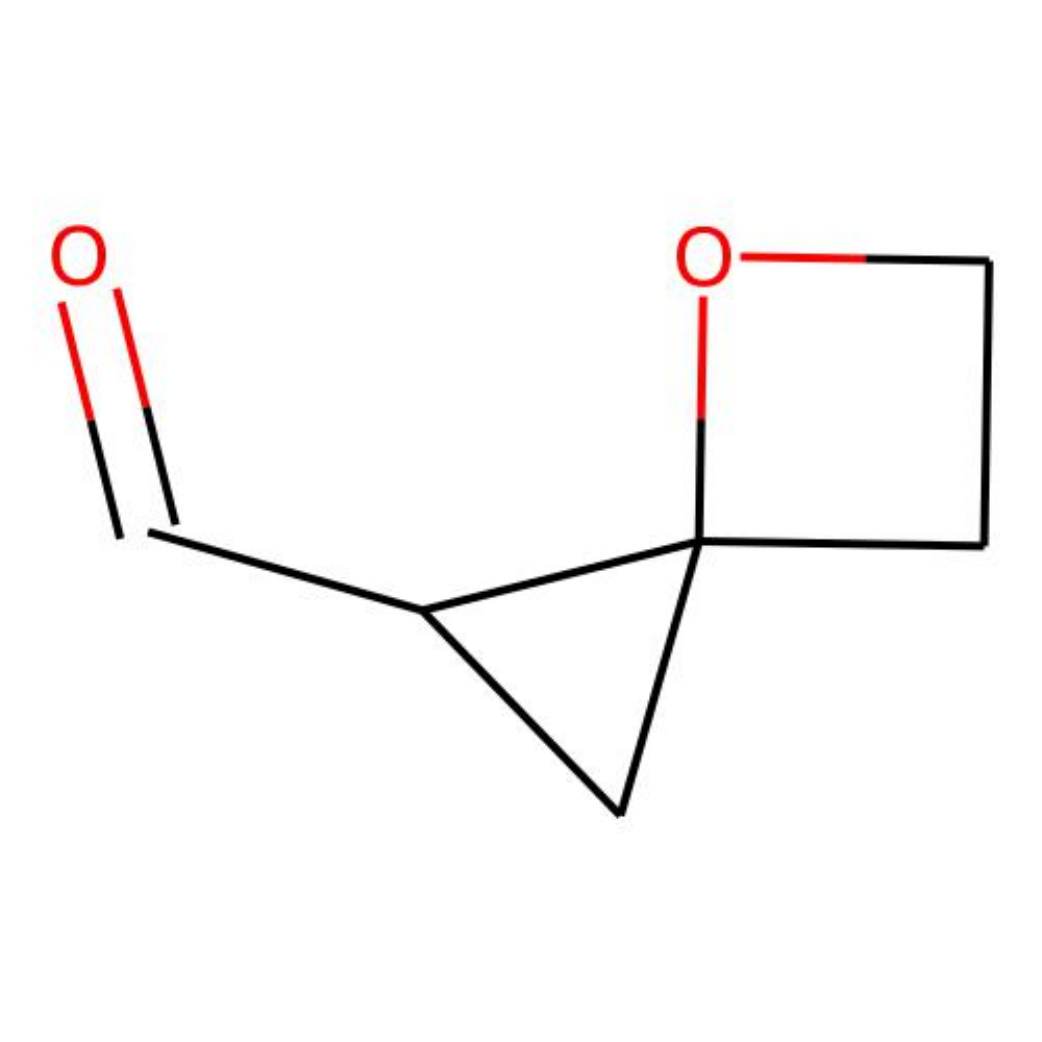}} &
        \multicolumn{1}{X|}{\includegraphics[width=0.08\textwidth]{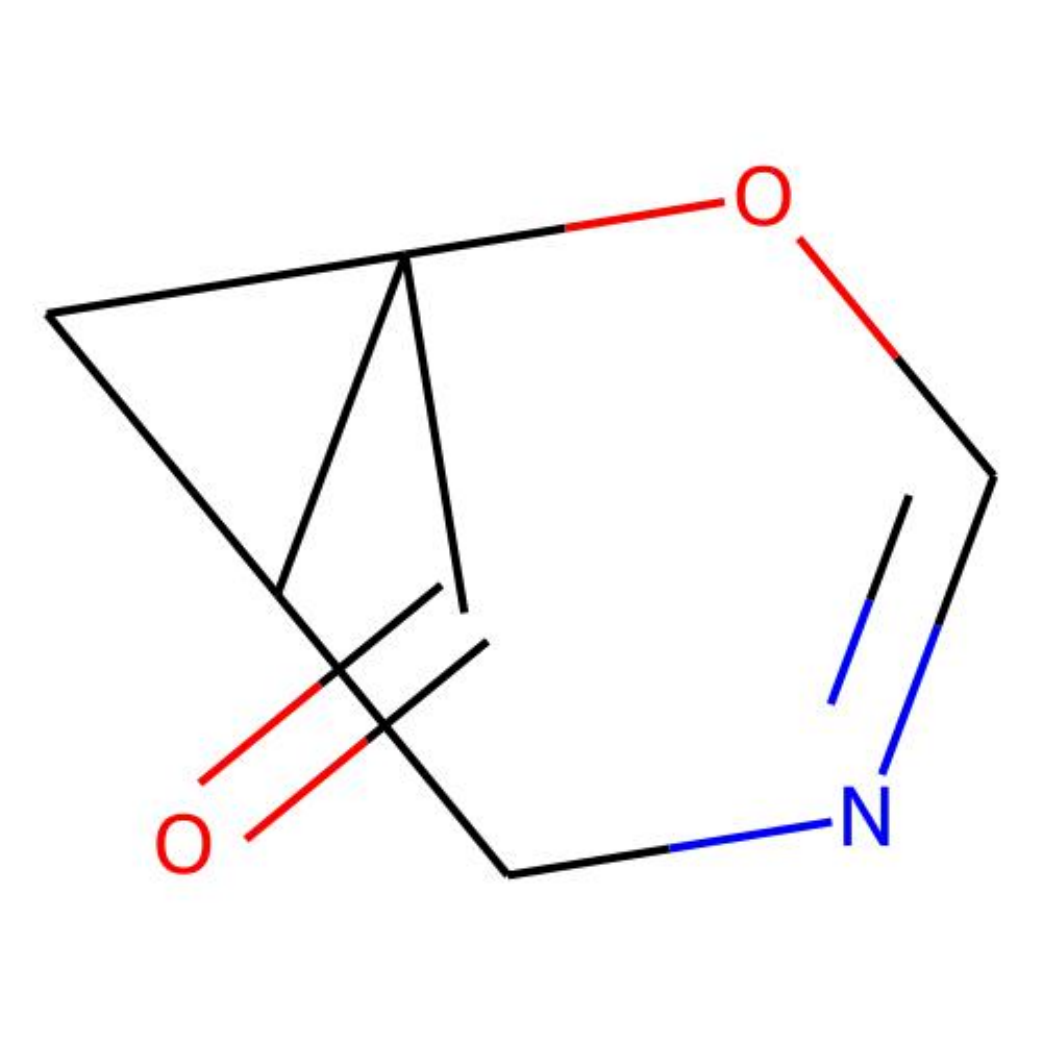}}&
        \multicolumn{1}{c|}{\includegraphics[width=0.08\textwidth]{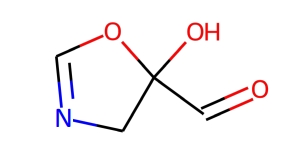}} &
        \multicolumn{1}{c|}{\includegraphics[width=0.08\textwidth]{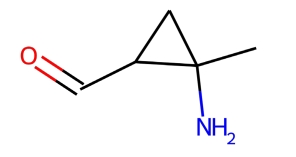}} &
        \multicolumn{1}{c|}{\includegraphics[width=0.08\textwidth]{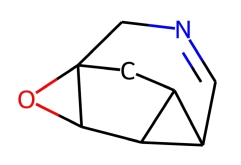}} &
        \multicolumn{1}{c|}{\includegraphics[width=0.08\textwidth]{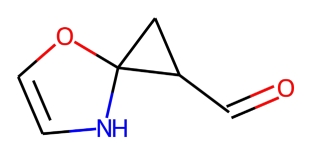}}&
        \multicolumn{1}{c|}{\includegraphics[width=0.08\textwidth]{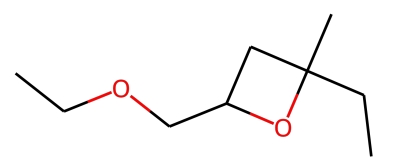}}&
        \multicolumn{1}{c}{\includegraphics[width=0.08\textwidth]{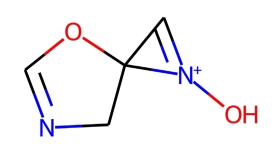}} \\
        similarity & \textbf{0.4516} & 0.4000 & 0.4242 & 
        0.3125 & 0.2750 & 0.3514
        & 0.1667 & 0.3529 \\
        \hline
        \multicolumn{1}{c|}{\includegraphics[width=0.08\textwidth]{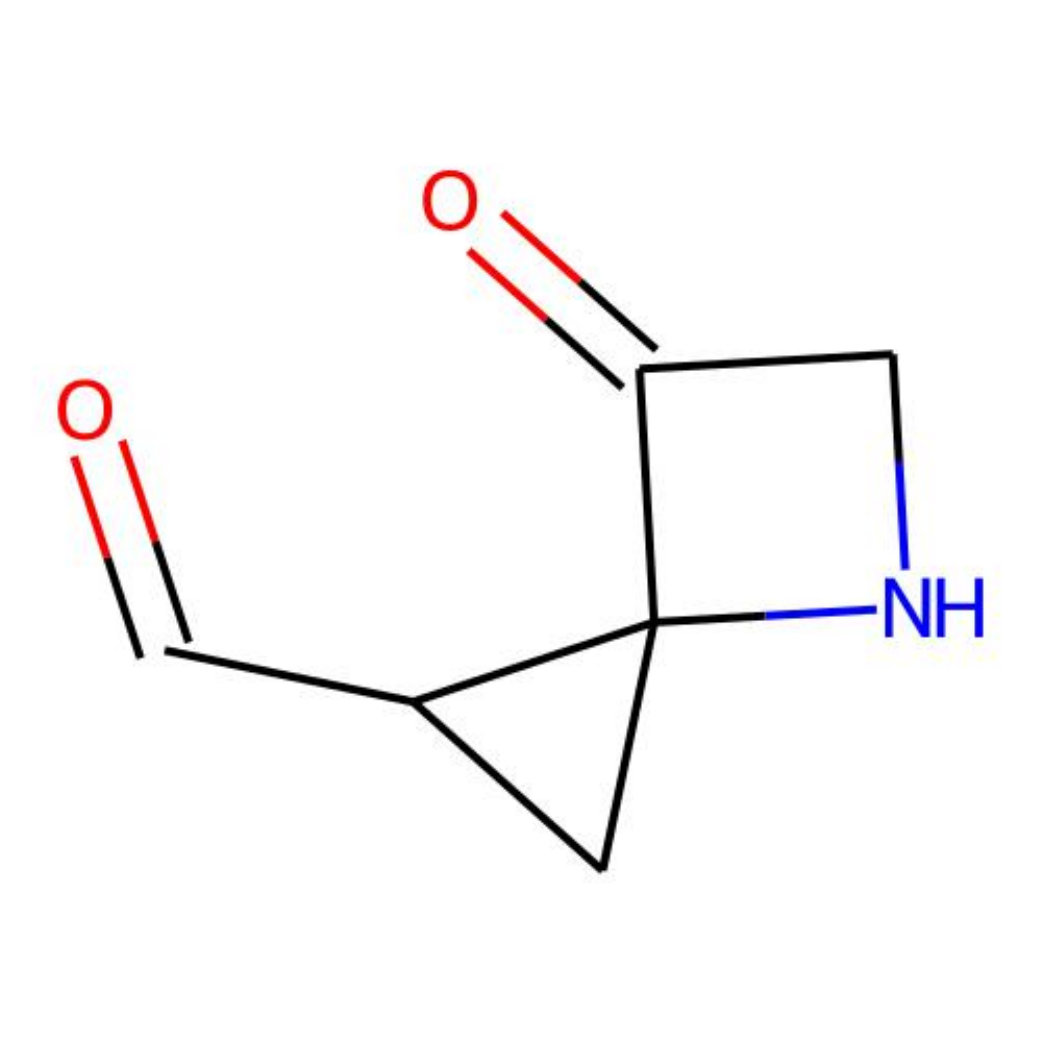}} &
        \multicolumn{1}{c|}{\includegraphics[width=0.08\textwidth]{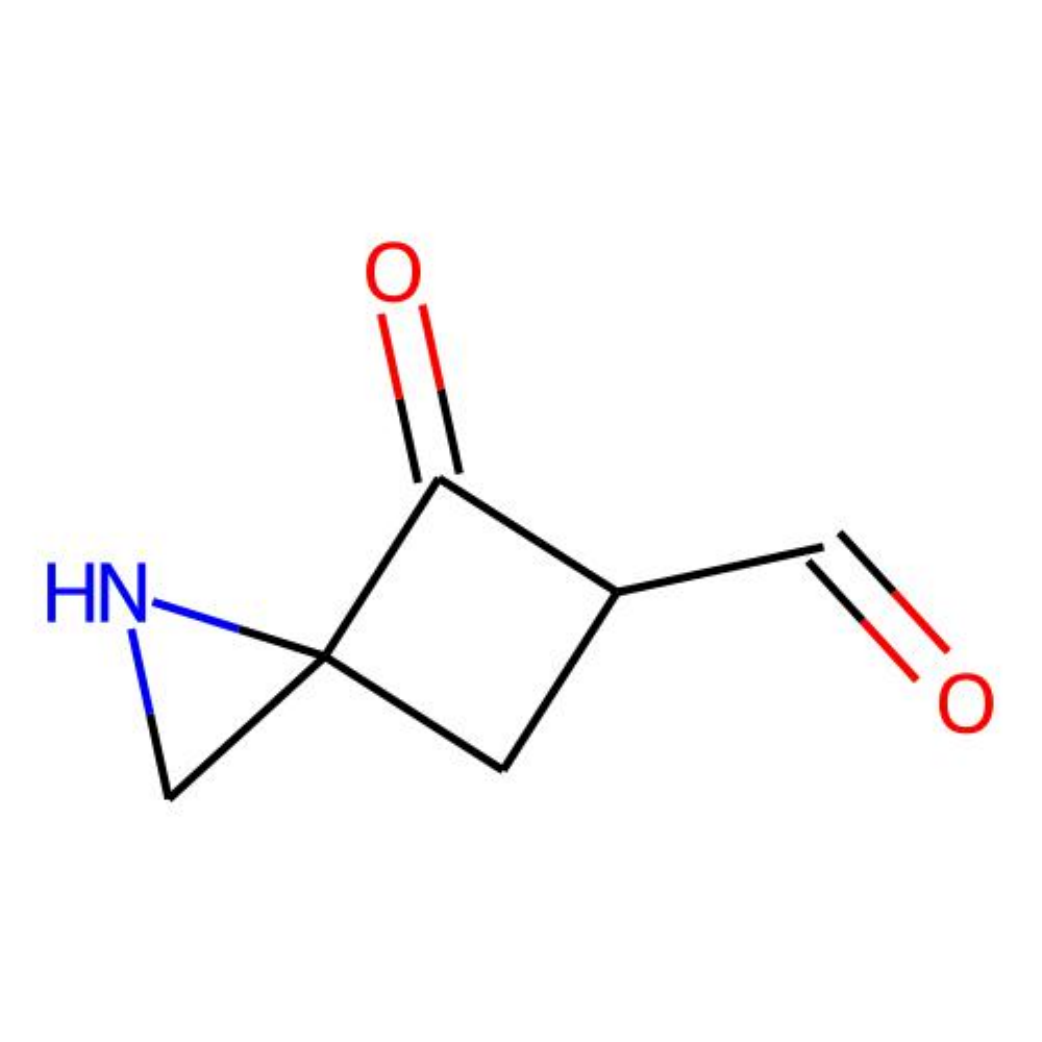}} &
        \multicolumn{1}{X|}{\includegraphics[width=0.08\textwidth]{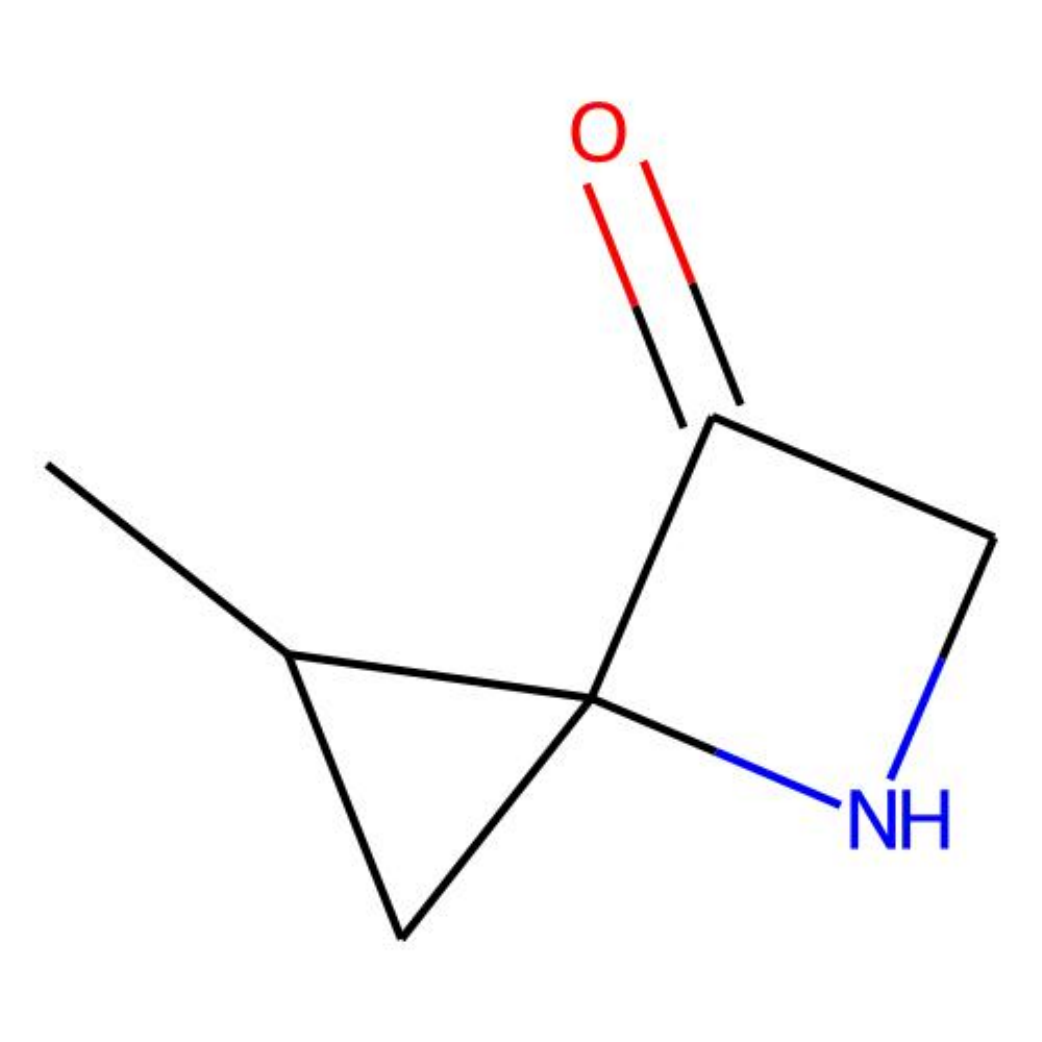}}&
        \multicolumn{1}{c|}{\includegraphics[width=0.08\textwidth]{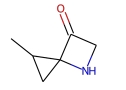}} &
        \multicolumn{1}{c|}{\includegraphics[width=0.08\textwidth]{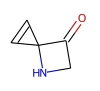}} &
        \multicolumn{1}{c|}{\includegraphics[width=0.08\textwidth]{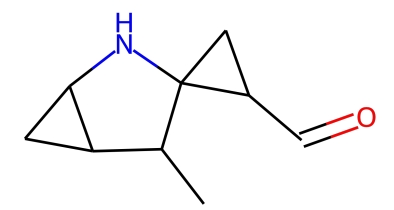}} &
        \multicolumn{1}{c|}{\includegraphics[width=0.08\textwidth]{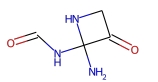}}&
        \multicolumn{1}{c|}{\includegraphics[width=0.08\textwidth]{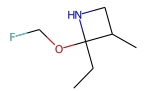}}&
        \multicolumn{1}{c}{\includegraphics[width=0.08\textwidth]{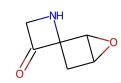}}\\
        similarity & 0.3636 & \textbf{0.5357} & \textbf{0.5357} &
        0.3333 &0.2821 & 0.3529 &
        0.1707 &0.4688 \\
       \hline
       \hline
        Zin250k Ref. & Ours & Grum& GDSS & GDSS-seq 
        & GraphAF & MoFlow & GraphDF & EDP-GNN  \\
        \hline
        \multicolumn{1}{c|}{\includegraphics[width=0.08\textwidth]{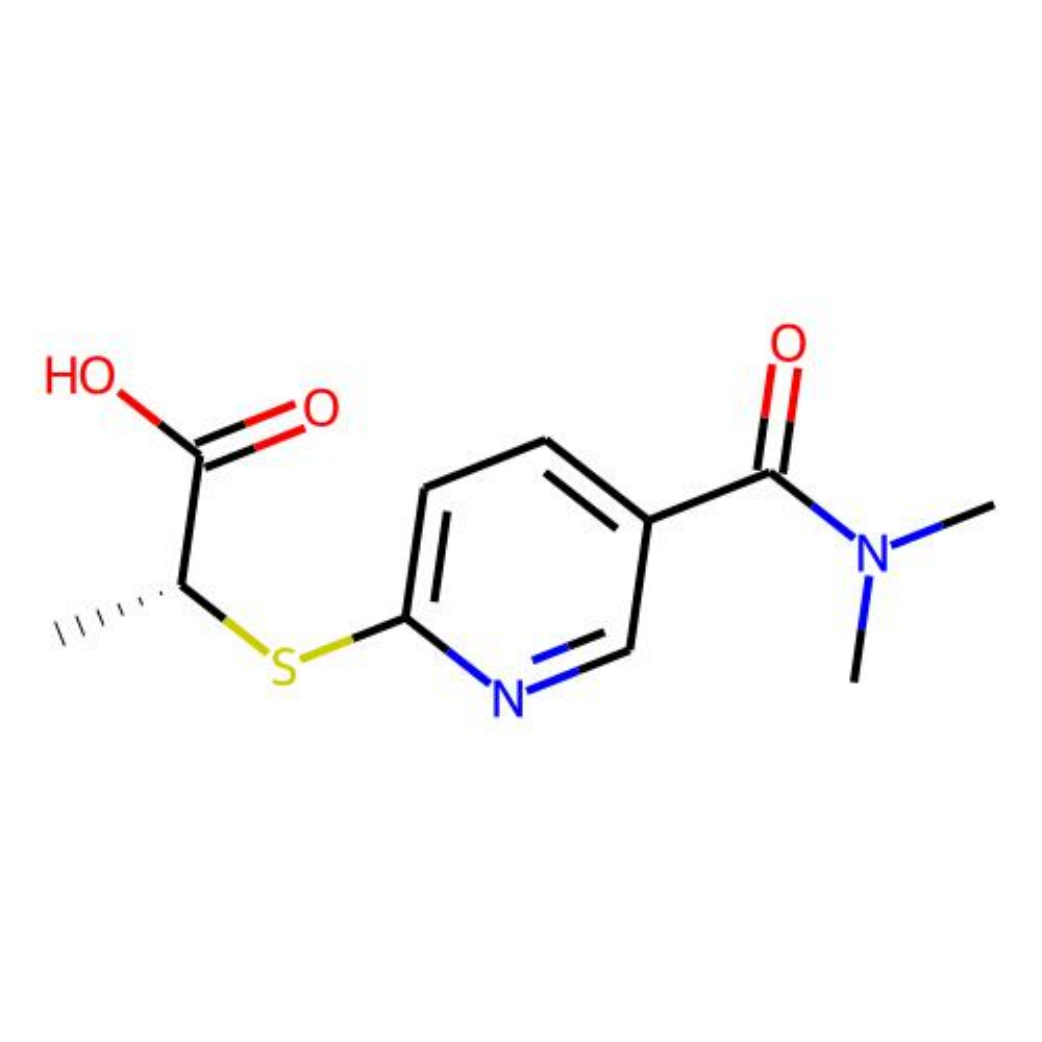}} &
        \multicolumn{1}{c|}{\includegraphics[width=0.08\textwidth]{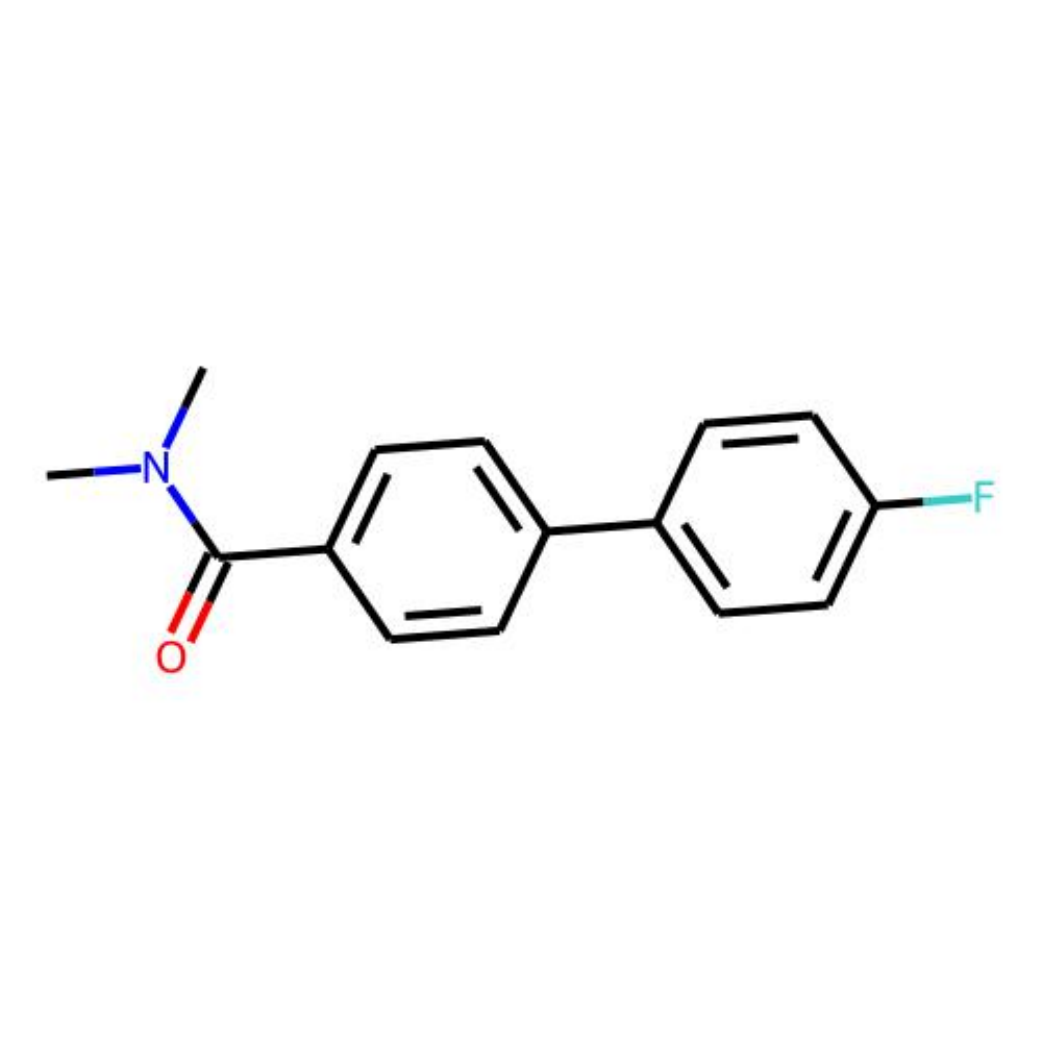}} &
        \multicolumn{1}{X|}{\includegraphics[width=0.08\textwidth]{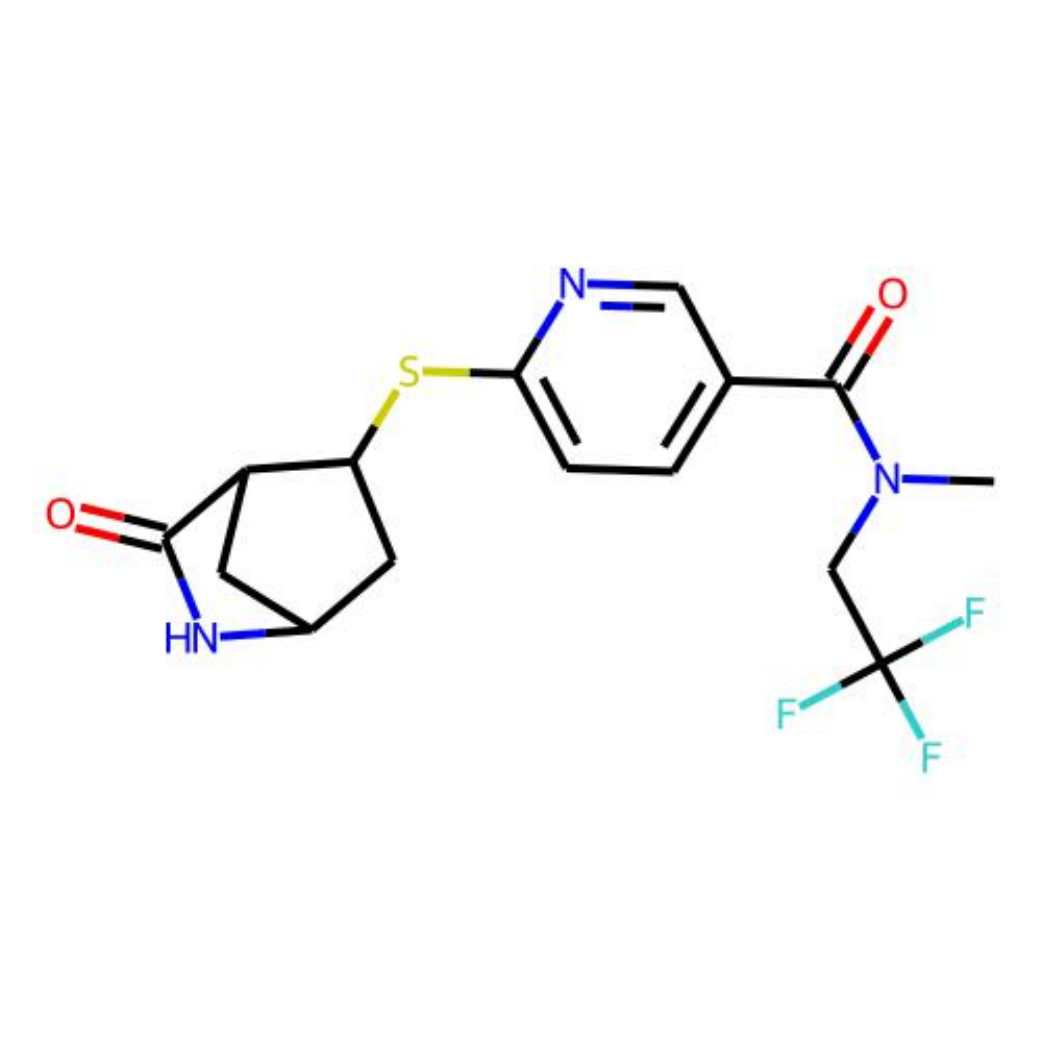}}&
        \multicolumn{1}{c|}{\includegraphics[width=0.08\textwidth]{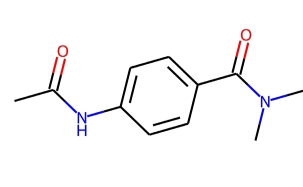}} &
        \multicolumn{1}{c|}{\includegraphics[width=0.08\textwidth]{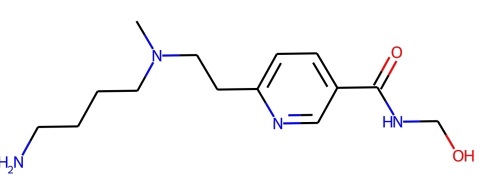}} &
        \multicolumn{1}{c|}{\includegraphics[width=0.08\textwidth]{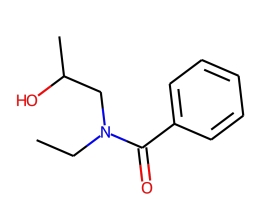}} &
        \multicolumn{1}{c|}{\includegraphics[width=0.08\textwidth]{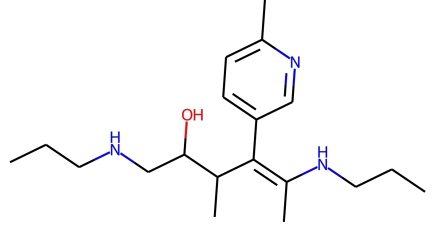}}&
        \multicolumn{1}{c|}{\includegraphics[width=0.08\textwidth]{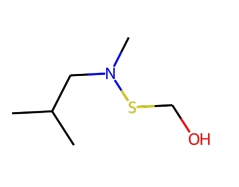}}&
        \multicolumn{1}{c}{\includegraphics[width=0.08\textwidth]{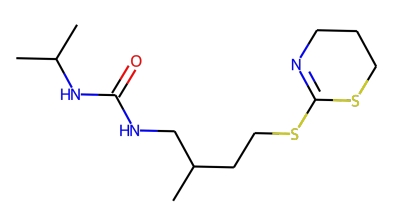}} \\
        similarity & \textbf{0.3809} & 0.3606 
        &0.3191 &0.2875 &0.2885 &0.2462 &0.2128 &0.2540
 \\
        \hline
        \multicolumn{1}{c|}{\includegraphics[width=0.08\textwidth]{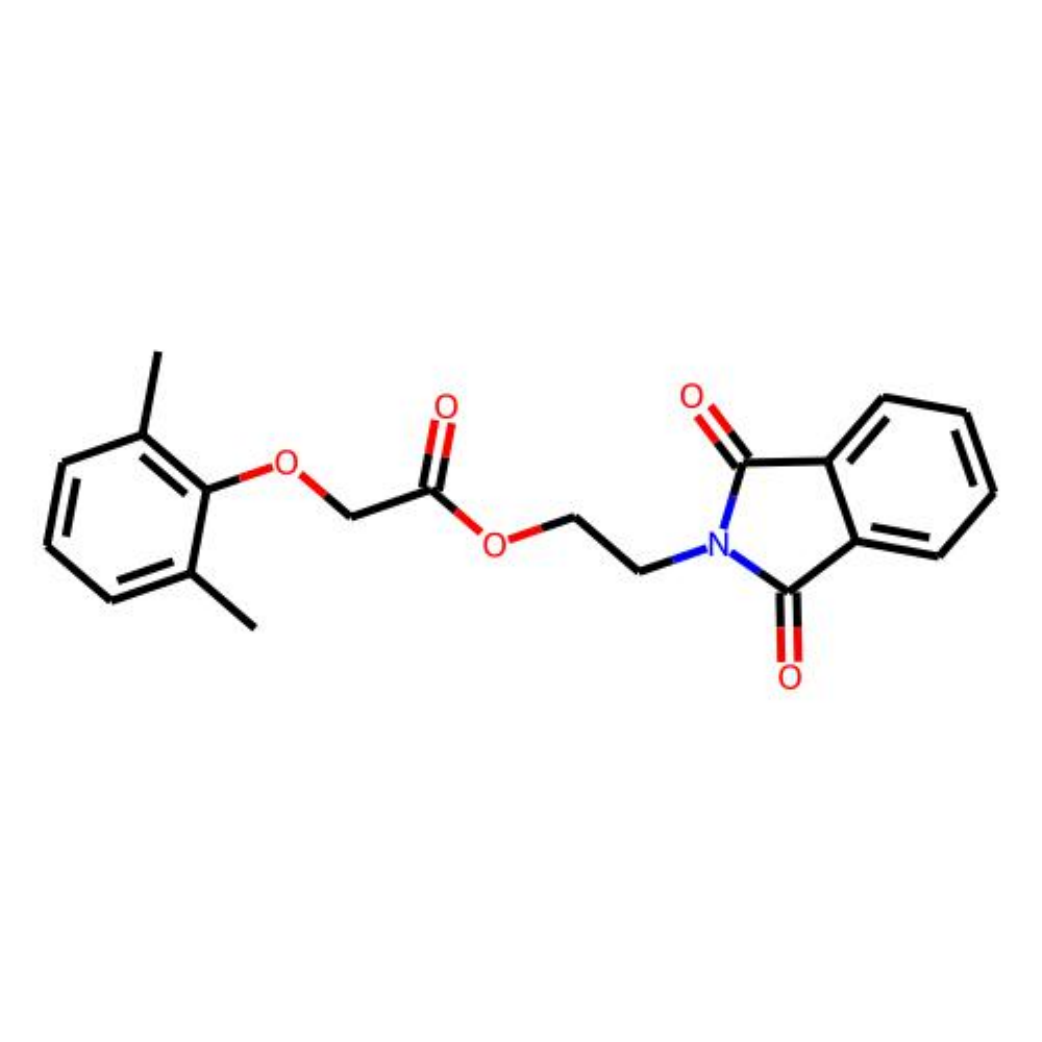}} &
        \multicolumn{1}{c|}{\includegraphics[width=0.08\textwidth]{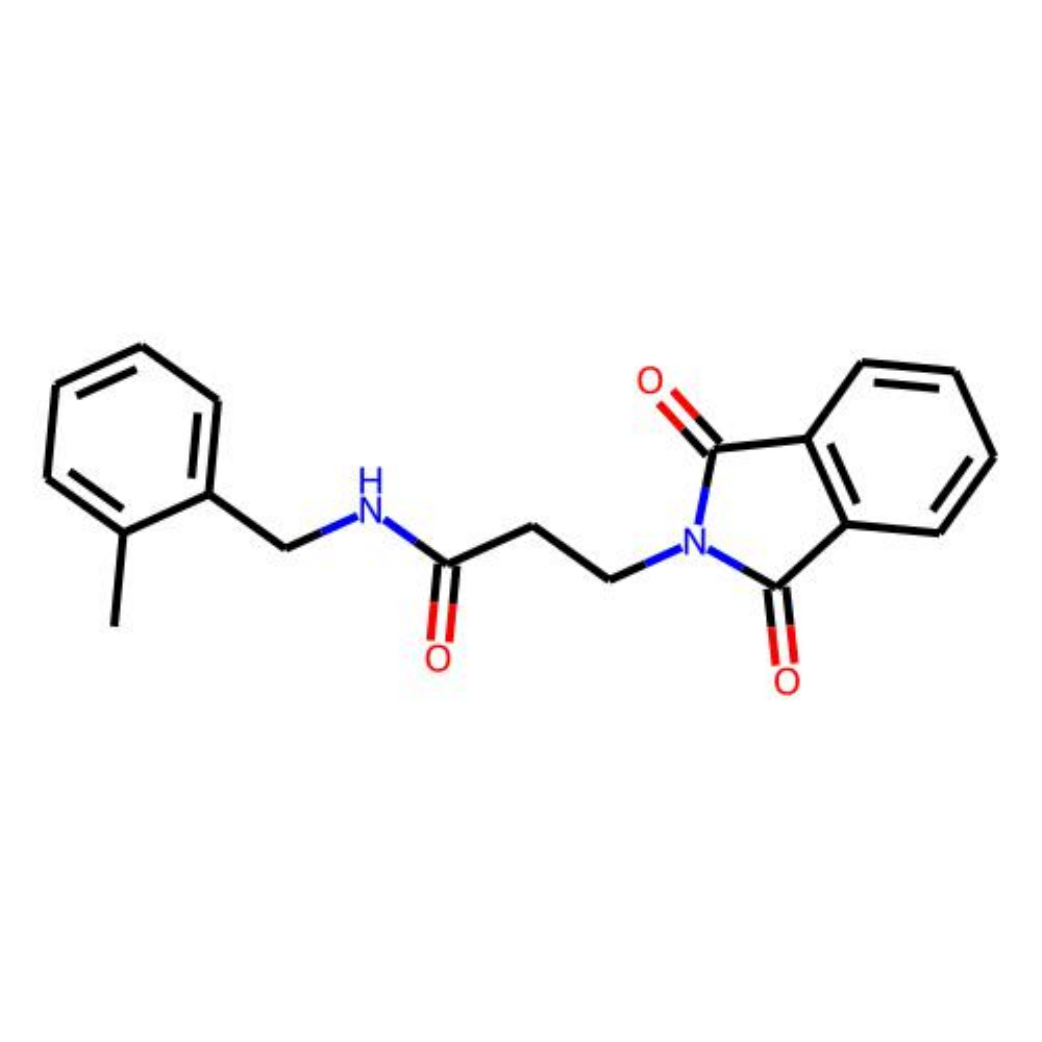}} &
        \multicolumn{1}{X|}{\includegraphics[width=0.08\textwidth]{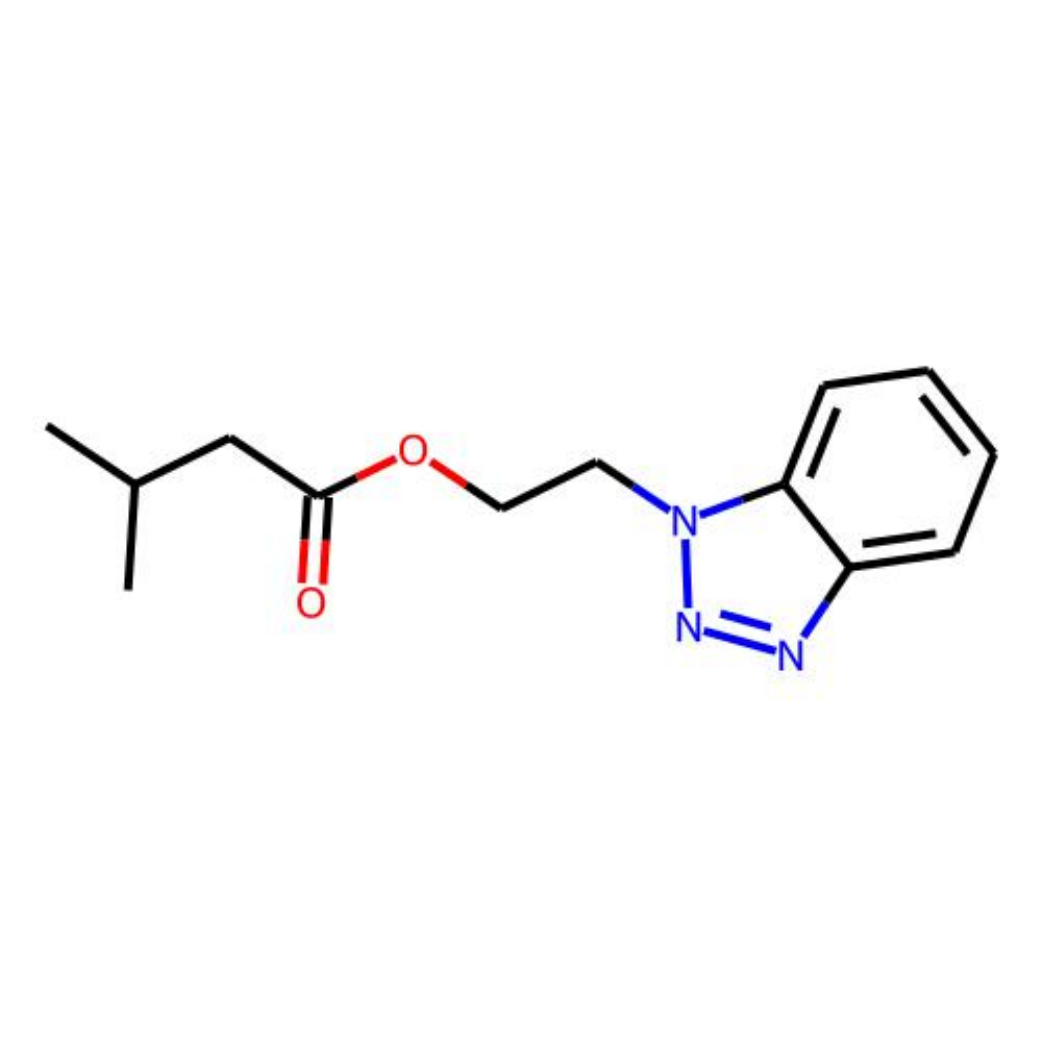}}&
        \multicolumn{1}{c|}{\includegraphics[width=0.08\textwidth]{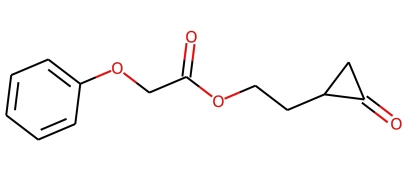}} &
        \multicolumn{1}{c|}{\includegraphics[width=0.08\textwidth]{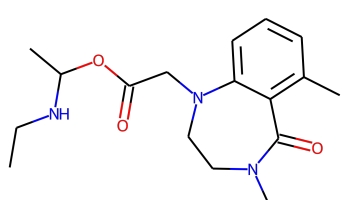}} &
        \multicolumn{1}{c|}{\includegraphics[width=0.08\textwidth]{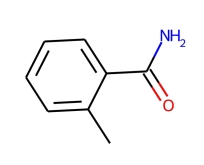}} &
        \multicolumn{1}{c|}{\includegraphics[width=0.08\textwidth]{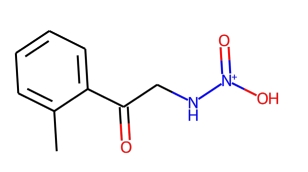}}&
        \multicolumn{1}{c|}{\includegraphics[width=0.08\textwidth]{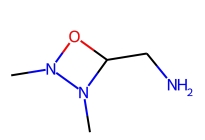}}&
        \multicolumn{1}{c}{\includegraphics[width=0.08\textwidth]{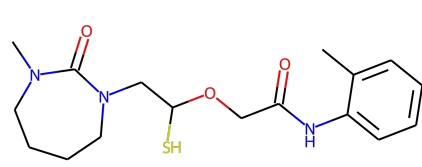}} \\
        similarity & \textbf{0.4615} & 0.3275  
        &0.3519 &0.2941 &0.3111 &0.2593 &0.1633 &0.2571 \\
        \hline
        \multicolumn{1}{c|}{\includegraphics[width=0.08\textwidth]{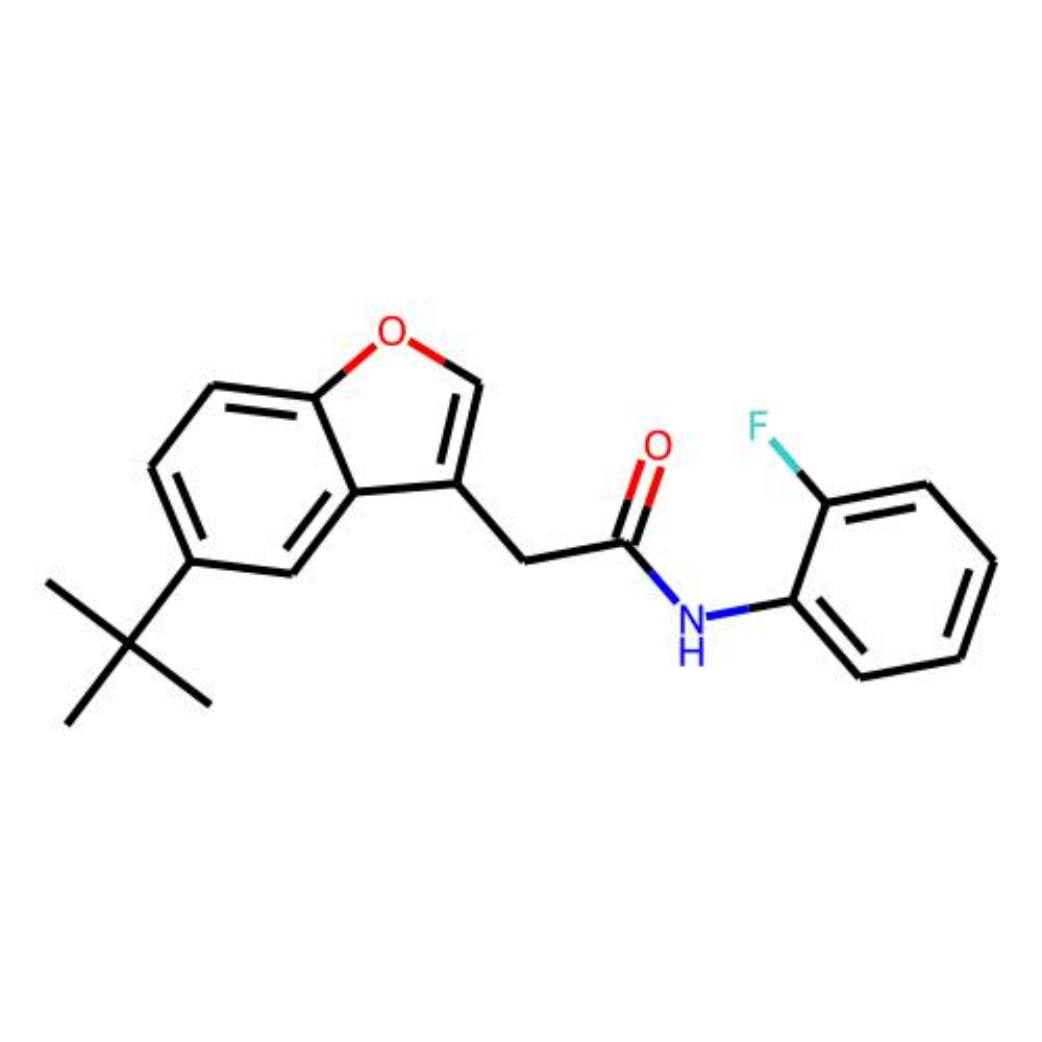}} &
        \multicolumn{1}{c|}{\includegraphics[width=0.08\textwidth]{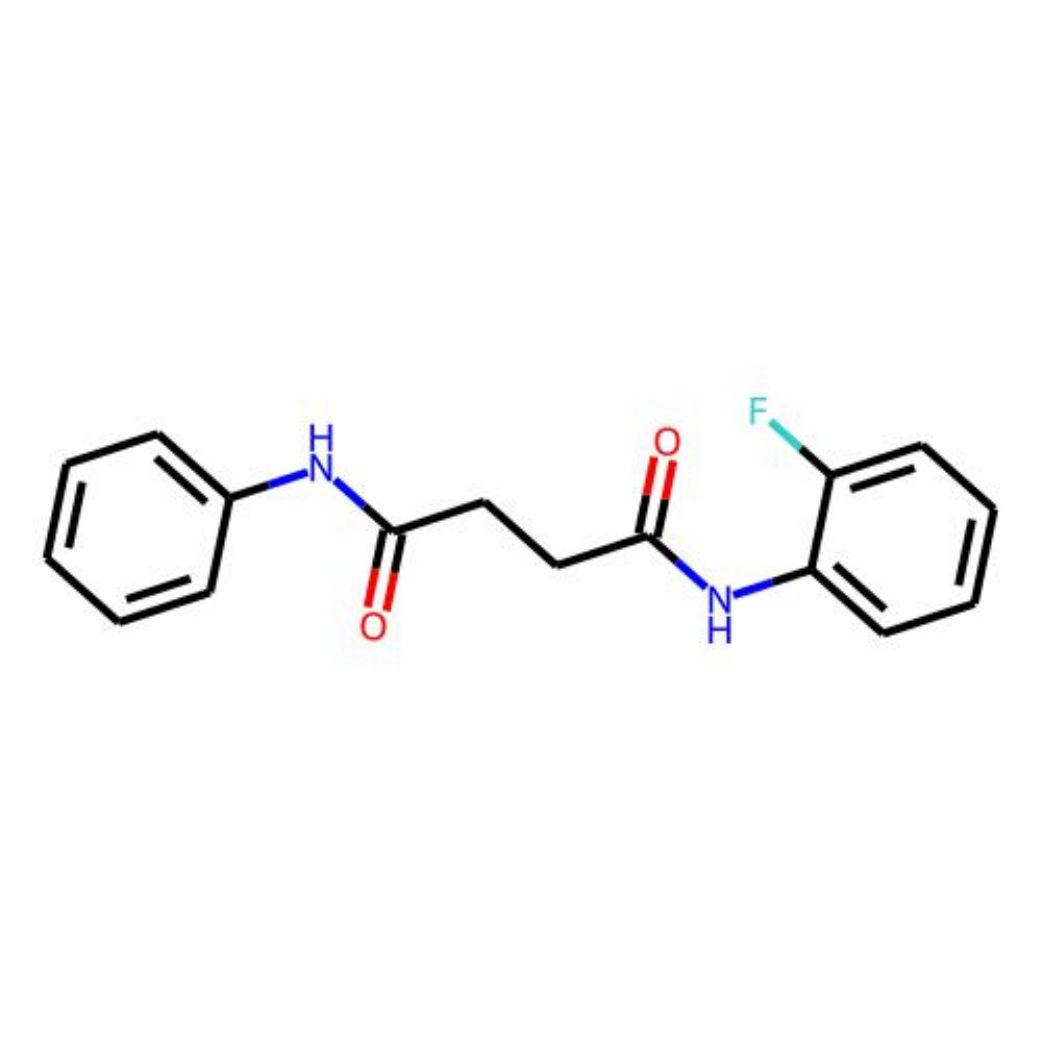}} &
        \multicolumn{1}{X|}{\includegraphics[width=0.08\textwidth]{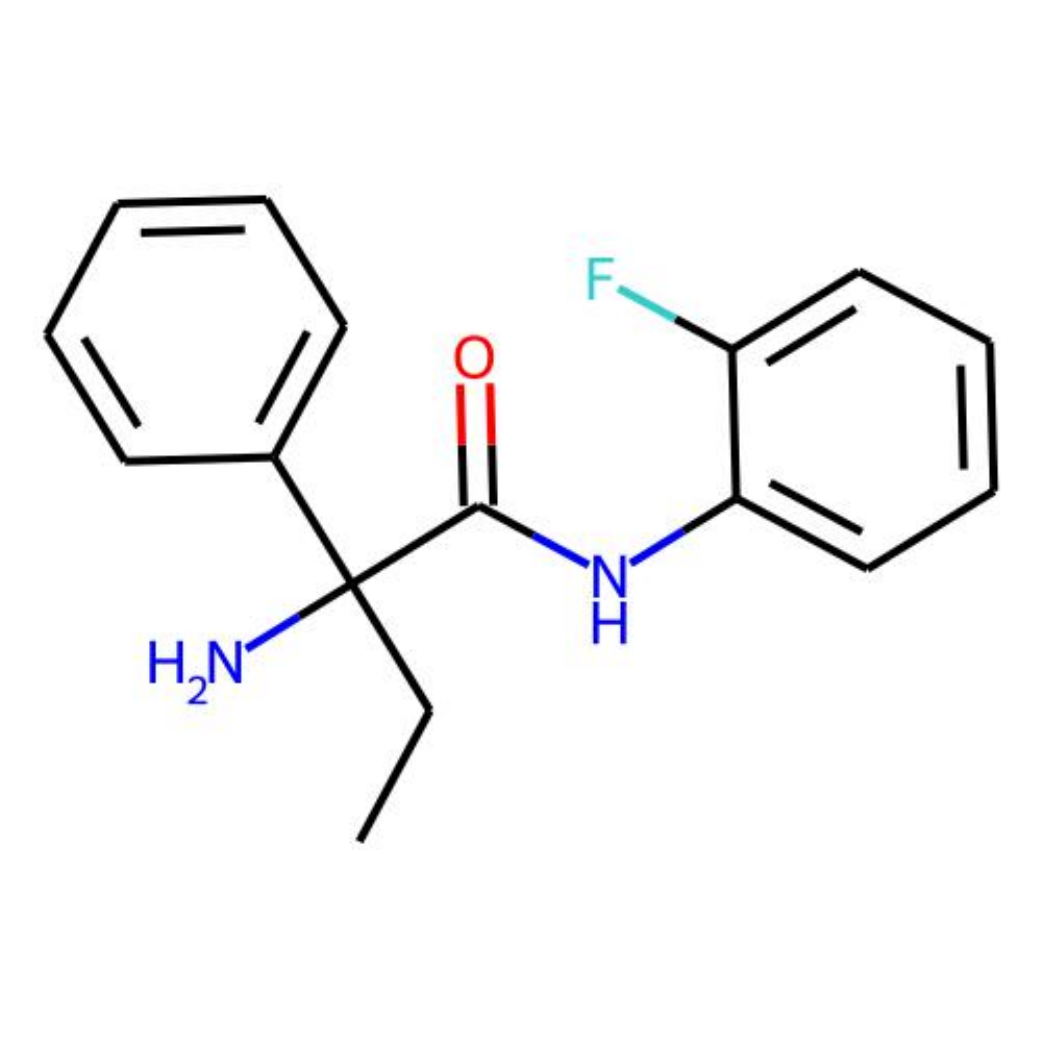}}&
        \multicolumn{1}{c|}{\includegraphics[width=0.08\textwidth]{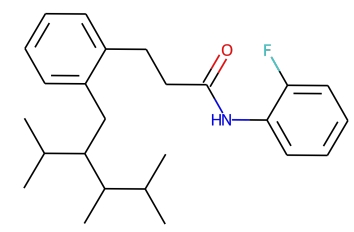}} &
        \multicolumn{1}{c|}{\includegraphics[width=0.08\textwidth]{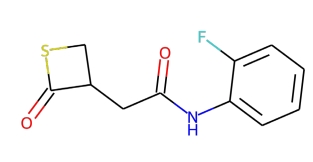}} &
        \multicolumn{1}{c|}{\includegraphics[width=0.08\textwidth]{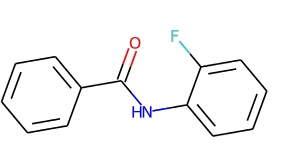}} &
        \multicolumn{1}{c|}{\includegraphics[width=0.08\textwidth]{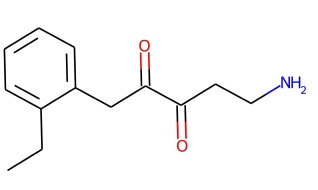}}&
        \multicolumn{1}{c|}{\includegraphics[width=0.08\textwidth]{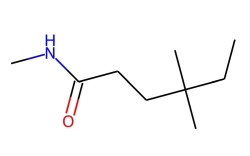}}&
        \multicolumn{1}{c}{\includegraphics[width=0.08\textwidth]{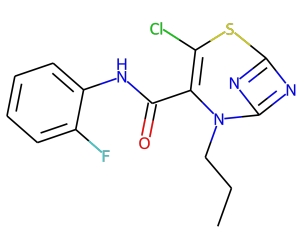}} \\
        similarity & \textbf{0.4038} & 0.3859
        &0.3871 &0.3559 &0.3800 
        &0.2143 &0.1786 &0.3043\\
        \hline
    \end{tabularx}
    \caption{Visualization of the generated molecules with Tanimoto similarity scores based on Morgan fingerprints. The best results are highlighted in bold.}
    \label{fig:visu}
\end{figure*}

Table~\ref{tab:exp} presents performance comparisons on both the QM9 and ZINC250k datasets against baseline models. Our approach consistently achieves top-two validity scores across both datasets, demonstrating its effectiveness in enabling the LLM to capture the underlying chemical rules essential for accurate molecule generation. For novelty, our method attains a perfect score of 100\% on the ZINC250k dataset and 88\% on QM9, highlighting its ability to consistently generate novel molecular structures.
In terms of FCD and Scaf metrics—critical indicators of a model's ability to explore and replicate chemical space—our method delivers competitive performance compared to other baselines. While DiGress and Grum show strong FCD and Scaf scores on the QM9 dataset, their novelty scores fall significantly short (below 40\%), suggesting potential overfitting to the training data rather than true generalization of molecular distributions. In contrast, our method not only maintains high novelty rates but also achieves strong performance on FCD and Scaf metrics.
On the ZINC250k dataset, our approach attains the highest Scaf score and the second-best FCD score, further demonstrating its superior ability to generalize and innovate within chemical spaces. This robust performance underscores our model’s advanced understanding and application of molecular distributions, making it a powerful tool for innovative molecular design in computational chemistry.

\subsection{Visualization Results of Generated Molecules}

In Fig.~\ref{fig:visu}, we follow the experimental setup outlined in \citep{jo2022score}, using Tanimoto similarity based on Morgan fingerprints to evaluate the generated molecular graphs. For consistency and comparability, we select the same molecules as \citep{jo2022score}. Additionally, we perform experiments on molecular graphs generated by Grum \citep{jo2023graph}. Across most cases, our method demonstrates superior performance compared to previous state-of-the-art diffusion-based approaches, showcasing its effectiveness and robustness in molecular graph generation.

\subsection{Ablation Study: Impact of Tree-Structured Text Encoding}

\begin{wraptable}{r}{0.6\textwidth}
    \vspace{-15pt}
    \small
    \centering
    \caption{Study of the impact of tree-structured text encoding on the ZINC250K dataset.}
    \label{tab:ablation_enc_zin250}
    \begin{tabularx}{\linewidth}{l|YYYY}
    \toprule
    {Methods}  & {Valid}$\uparrow$ & {FCD}$\downarrow$ & {Scaf}$\uparrow$ & {Novelty}$\uparrow$ \\ \midrule

    Talk like a graph   &  59.20 & 19.8114  
    & 0.1317  & 100 \\
    Ours     & 98.60 & 5.6906
    & 0.1522  & 100\\ \bottomrule
    \end{tabularx}
\end{wraptable}
To evaluate how our proposed graph-to-tree text encoding improves the LLM's ability to learn graph structures compared to the previous graph-to-text methods such as Talk Like a Graph \citep{fatemi2023talk}, we conducted experiments on the challenging Zinc250K dataset \citep{irwin2012zinc}, which contains larger molecules. Talk Like a Graph encodes graph structures by converting them into natural language, where each node’s connections and attributes are described in sentence form.
For the fine-tuning process, we randomly selected 5,000 molecules from the training set and generated 1,000 molecules for performance comparison. As shown in Table~\ref{tab:ablation_enc_zin250}, our method significantly outperforms the previous approach across all metrics, demonstrating that encoding molecular structures in JSON format enables LLMs to more effectively learn and replicate complex molecular structures.

\subsection{Ablation Study: Impact of supervised Fine-Tuning LLM}
\begin{wraptable}{r}{0.5\textwidth}
    \vspace{-5pt}
    \centering
    \caption{Comparison of LLM performance with and without SFT on the ZINC250k dataset.}
    \label{tab:abl_sft}
    \begin{tabularx}{\linewidth}{l|YYY}
        \toprule
        Methods & Valid$\uparrow$  & Unique$\uparrow$ & Novelty$\uparrow$  \\
        \midrule
        w/o SFT & 70.80 & 61.12 & 100.00  \\
        w/ SFT & 98.60 & 98.98 & 100.00 \\
        \bottomrule
    \end{tabularx}
\end{wraptable}

In this study, we aim to evaluate the impact of supervised fine-tuning on LLM performance. Specifically, we generate 1,000 molecules using the same prompt to compare the performance of the LLM before and after fine-tuning. This direct comparison allows us to assess how fine-tuning enhances the model's ability to accurately generate molecular structures. We conduct this experiment using the ZINC250k dataset, and the results are presented in Table~\ref{tab:abl_sft}. The results reveal that without fine-tuning, the LLM produces molecules with only 70.8\% validity and 61.12\% uniqueness, indicating that the model, in its initial state, struggles to fully comprehend and accurately replicate the text representation of molecular structures. However, after fine-tuning, there is a significant improvement, with validity and uniqueness increasing to 99.6\% and 99.79\%, respectively. These results highlight the effectiveness of fine-tuning in substantially improving the model's performance, demonstrating its critical role in enabling the LLM to better understand and generate precise molecular structures.

\subsection{Ablation Study: Impact of size of the Fine-Tuning dataset}

\begin{wraptable}{r}{0.55\textwidth}
    \vspace{-15pt}
    \centering
    \caption{Comparison of LLM performance with different size of fine-tuning datasets}
    \label{tab:ablation_size}
    \begin{tabularx}{\linewidth}{l|YYYY}
    \toprule
    {Methods}  & {Valid}$\uparrow$ & {Novelty}$\uparrow$ & {FCD} $\downarrow$ & {Scaf} $\uparrow$   \\ \midrule

    1k (10 epoch)     & 98.50  & 90.38 & 1.226 & 0.6933 \\
    5k (10 epoch)      & 98.70 & 86.53  & 1.219 & 0.7779\\
    10k (10 epoch)     & 98.50  & 73.89 & 1.146 & 0.7980 \\ \bottomrule
    \end{tabularx}
    \vspace{-10pt}
\end{wraptable}
In this section, we investigate the impact of dataset size on the performance of a LLM during fine-tuning. Our experiments use the QM9 dataset with three distinct dataset sizes for fine-tuning: 1,000, 5,000, and 10,000 molecules. Each model is trained over 10 epochs. This setup enables a systematic evaluation of how variations in fine-tuning data size affect the model's learning efficacy and its ability to generalize. Table~\ref{tab:ablation_size} presents the results of these experiments.
The results indicate an improvement in the FCD and Scaf scores as the dataset size increases. This improvement likely stems from the LLM's exposure to a larger array of data points, which enhances its understanding of the chemical distribution within the dataset. Conversely, we observe a decrease in novelty scores with larger datasets. This reduction may be attributed to the relatively small and structurally simple nature of the QM9 dataset, which comprises only four types of atoms and molecules not exceeding nine atoms. As the model encounters more data, it increasingly reproduces similar outputs, reflecting the limited diversity in the dataset.

\subsection{Ablation Study: Impact of Token Constraining}
\begin{wraptable}{r}{0.5\textwidth}
    \vspace{-15pt}
    \centering
    \caption{Comparison results of using token constraining (TC) on molecular generation on the ZINC250k dataset.}
    \label{tab:abl_token}
    \begin{tabularx}{\linewidth}{l|YY}
        \toprule
                & w/o TC & w/ TC  \\\midrule
        Validity (\%)$\uparrow$  & 41.60 & 98.60 \\
        \bottomrule
    \end{tabularx}
\end{wraptable}

In this section, we examine the impact of token constraining on molecular generation, as introduced in Section~\ref{sec:3.3}. Token constraining is implemented to guide the LLM toward generating valid molecular structures by restricting its output to adhere to chemical rules. To evaluate the effectiveness of this approach, we perform an experimental comparison using the ZINC250k dataset. Specifically, we generate 1,000 molecules to compare the validity of the output with and without token constraining. The results, presented in Table~\ref{tab:abl_token}, clearly demonstrate the efficacy of token constraining in improving the validity of generated molecules. Without token constraining, the validity of the generated molecules is only 41.6\%. However, when token constraining is applied, validity dramatically increases to 98.6\%. This significant improvement underscores the critical role of token constraining in guiding the LLM to produce valid molecular structures, ensuring closer adherence to the fundamental rules of chemical structure and leading to a higher rate of valid outputs.

\section{Conclusion}

In this work, we introduced G2T-LLM, a novel approach for molecular generation that leverages LLMs to generate valid molecular structures through a novel graph-to-tree text encoding. By converting molecular graphs into hierarchical representations inspired by SMILES but adapted for LLMs, we bridge the gap between non-linear molecular structures and sequential data processing. This encoding allows the LLM to understand the molecular structure better and produce coherent chemical outputs.
Our method addresses the challenges of generating valid molecular structures by introducing token constraints during the generation process, ensuring that the outputs respect some chemical and structural rules. Through supervised fine-tuning, we further align the LLM with molecular generation tasks, improving its ability to produce chemically valid molecules based on the learned data patterns from benchmark datasets like Zinc250K and QM9.
Our results demonstrate the effectiveness of G2T-LLM, achieving state-of-the-art performance on benchmark datasets. This work highlights the potential of utilizing LLMs in molecular design, opening up new avenues for AI-driven discoveries in chemistry. The combination of hierarchical encoding, token constraining, and fine-tuning proves to be a powerful strategy for tackling the complexities of molecular generation. Future work will focus on refining these techniques to enhance efficiency and explore further applications in drug discovery and material science.


\bibliography{iclr2025_conference}

\begin{thebibliography}{33}
\providecommand{\natexlab}[1]{#1}
\providecommand{\url}[1]{\texttt{#1}}
\expandafter\ifx\csname urlstyle\endcsname\relax
  \providecommand{\doi}[1]{doi: #1}\else
  \providecommand{\doi}{doi: \begingroup \urlstyle{rm}\Url}\fi

\bibitem[Ansel et~al.(2024)Ansel, Yang, He, Gimelshein, Jain, Voznesensky, Bao, Bell, Berard, Burovski, Chauhan, Chourdia, Constable, Desmaison, DeVito, Ellison, Feng, Gong, Gschwind, Hirsh, Huang, Kalambarkar, Kirsch, Lazos, Lezcano, Liang, Liang, Lu, Luk, Maher, Pan, Puhrsch, Reso, Saroufim, Siraichi, Suk, Suo, Tillet, Wang, Wang, Wen, Zhang, Zhao, Zhou, Zou, Mathews, Chanan, Wu, and Chintala]{Ansel_PyTorch_2_Faster_2024}
Jason Ansel, Edward Yang, Horace He, Natalia Gimelshein, Animesh Jain, Michael Voznesensky, Bin Bao, Peter Bell, David Berard, Evgeni Burovski, Geeta Chauhan, Anjali Chourdia, Will Constable, Alban Desmaison, Zachary DeVito, Elias Ellison, Will Feng, Jiong Gong, Michael Gschwind, Brian Hirsh, Sherlock Huang, Kshiteej Kalambarkar, Laurent Kirsch, Michael Lazos, Mario Lezcano, Yanbo Liang, Jason Liang, Yinghai Lu, CK~Luk, Bert Maher, Yunjie Pan, Christian Puhrsch, Matthias Reso, Mark Saroufim, Marcos~Yukio Siraichi, Helen Suk, Michael Suo, Phil Tillet, Eikan Wang, Xiaodong Wang, William Wen, Shunting Zhang, Xu~Zhao, Keren Zhou, Richard Zou, Ajit Mathews, Gregory Chanan, Peng Wu, and Soumith Chintala.
\newblock {PyTorch 2: Faster Machine Learning Through Dynamic Python Bytecode Transformation and Graph Compilation}.
\newblock In \emph{29th ACM International Conference on Architectural Support for Programming Languages and Operating Systems, Volume 2 (ASPLOS '24)}. ACM, April 2024.
\newblock \doi{10.1145/3620665.3640366}.
\newblock URL \url{https://pytorch.org/assets/pytorch2-2.pdf}.

\bibitem[Brahmavar et~al.(2024)Brahmavar, Srinivasan, Dash, Krishnan, Vig, Roy, and Aduri]{brahmavar2024generating}
Shreyas~Bhat Brahmavar, Ashwin Srinivasan, Tirtharaj Dash, Sowmya~Ramaswamy Krishnan, Lovekesh Vig, Arijit Roy, and Raviprasad Aduri.
\newblock Generating novel leads for drug discovery using llms with logical feedback.
\newblock In \emph{Proceedings of the AAAI Conference on Artificial Intelligence}, volume~38, pp.\  21--29, 2024.

\bibitem[Brown(2020)]{brown2020language}
Tom~B Brown.
\newblock Language models are few-shot learners.
\newblock \emph{arXiv preprint arXiv:2005.14165}, 2020.

\bibitem[Costa \& Grave(2010)Costa and Grave]{costa2010fast}
Fabrizio Costa and Kurt~De Grave.
\newblock Fast neighborhood subgraph pairwise distance kernel.
\newblock In \emph{Proceedings of the 27th International Conference on International Conference on Machine Learning}, pp.\  255--262, 2010.

\bibitem[Dai et~al.(2018)Dai, Tian, Dai, Skiena, and Song]{dai2018syntax}
Hanjun Dai, Yingtao Tian, Bo~Dai, Steven Skiena, and Le~Song.
\newblock Syntax-directed variational autoencoder for structured data.
\newblock \emph{arXiv preprint arXiv:1802.08786}, 2018.

\bibitem[Dettmers et~al.(2024)Dettmers, Pagnoni, Holtzman, and Zettlemoyer]{dettmers2024qlora}
Tim Dettmers, Artidoro Pagnoni, Ari Holtzman, and Luke Zettlemoyer.
\newblock Qlora: Efficient finetuning of quantized llms.
\newblock \emph{Advances in Neural Information Processing Systems}, 36, 2024.

\bibitem[Dubey et~al.(2024)Dubey, Jauhri, Pandey, Kadian, Al-Dahle, Letman, Mathur, Schelten, Yang, Fan, et~al.]{dubey2024llama}
Abhimanyu Dubey, Abhinav Jauhri, Abhinav Pandey, Abhishek Kadian, Ahmad Al-Dahle, Aiesha Letman, Akhil Mathur, Alan Schelten, Amy Yang, Angela Fan, et~al.
\newblock The llama 3 herd of models.
\newblock \emph{arXiv preprint arXiv:2407.21783}, 2024.

\bibitem[Elton et~al.(2019)Elton, Boukouvalas, Fuge, and Chung]{113}
Daniel~C Elton, Zois Boukouvalas, Mark~D Fuge, and Peter~W Chung.
\newblock Deep learning for molecular design—a review of the state of the art.
\newblock \emph{Molecular Systems Design \& Engineering}, 4\penalty0 (4):\penalty0 828--849, 2019.

\bibitem[Fatemi et~al.(2023)Fatemi, Halcrow, and Perozzi]{fatemi2023talk}
Bahare Fatemi, Jonathan Halcrow, and Bryan Perozzi.
\newblock Talk like a graph: Encoding graphs for large language models.
\newblock \emph{arXiv preprint arXiv:2310.04560}, 2023.

\bibitem[Irwin et~al.(2012)Irwin, Sterling, Mysinger, Bolstad, and Coleman]{irwin2012zinc}
John~J Irwin, Teague Sterling, Michael~M Mysinger, Erin~S Bolstad, and Ryan~G Coleman.
\newblock Zinc: a free tool to discover chemistry for biology.
\newblock \emph{Journal of chemical information and modeling}, 52\penalty0 (7):\penalty0 1757--1768, 2012.

\bibitem[Jo et~al.(2022)Jo, Lee, and Hwang]{jo2022score}
Jaehyeong Jo, Seul Lee, and Sung~Ju Hwang.
\newblock Score-based generative modeling of graphs via the system of stochastic differential equations.
\newblock In \emph{International conference on machine learning}, pp.\  10362--10383. PMLR, 2022.

\bibitem[Jo et~al.(2023)Jo, Kim, and Hwang]{jo2023graph}
Jaehyeong Jo, Dongki Kim, and Sung~Ju Hwang.
\newblock Graph generation with diffusion mixture.
\newblock \emph{arXiv preprint arXiv:2302.03596}, 2023.

\bibitem[Le et~al.(2024)Le, Guo, Dong, Huang, Nan, Iyer, Zhang, Wiest, Wang, and Chawla]{le2024molx}
Khiem Le, Zhichun Guo, Kaiwen Dong, Xiaobao Huang, Bozhao Nan, Roshni Iyer, Xiangliang Zhang, Olaf Wiest, Wei Wang, and Nitesh~V Chawla.
\newblock Molx: Enhancing large language models for molecular learning with a multi-modal extension.
\newblock \emph{arXiv preprint arXiv:2406.06777}, 2024.

\bibitem[Liu et~al.(2024)Liu, Ding, Zhou, Fan, and Tan]{liu2024moleculargpt}
Yuyan Liu, Sirui Ding, Sheng Zhou, Wenqi Fan, and Qiaoyu Tan.
\newblock Moleculargpt: Open large language model (llm) for few-shot molecular property prediction.
\newblock \emph{arXiv preprint arXiv:2406.12950}, 2024.

\bibitem[Loshchilov(2017)]{loshchilov2017decoupled}
I~Loshchilov.
\newblock Decoupled weight decay regularization.
\newblock \emph{arXiv preprint arXiv:1711.05101}, 2017.

\bibitem[Luo et~al.(2023)Luo, Yang, Hong, Liu, and Nie]{luo2023molfm}
Yizhen Luo, Kai Yang, Massimo Hong, Xing~Yi Liu, and Zaiqing Nie.
\newblock Molfm: A multimodal molecular foundation model.
\newblock \emph{arXiv preprint arXiv:2307.09484}, 2023.

\bibitem[Luo et~al.(2021)Luo, Yan, and Ji]{luo2021graphdf}
Youzhi Luo, Keqiang Yan, and Shuiwang Ji.
\newblock Graphdf: A discrete flow model for molecular graph generation.
\newblock In \emph{International conference on machine learning}, pp.\  7192--7203. PMLR, 2021.

\bibitem[Madhawa et~al.(2019)Madhawa, Ishiguro, Nakago, and Abe]{madhawa2019graphnvp}
Kaushalya Madhawa, Katushiko Ishiguro, Kosuke Nakago, and Motoki Abe.
\newblock Graphnvp: An invertible flow model for generating molecular graphs.
\newblock \emph{arXiv preprint arXiv:1905.11600}, 2019.

\bibitem[Niu et~al.(2020)Niu, Song, Song, Zhao, Grover, and Ermon]{niu2020permutation}
Chenhao Niu, Yang Song, Jiaming Song, Shengjia Zhao, Aditya Grover, and Stefano Ermon.
\newblock Permutation invariant graph generation via score-based generative modeling.
\newblock In \emph{International Conference on Artificial Intelligence and Statistics}, pp.\  4474--4484. PMLR, 2020.

\bibitem[Paszke et~al.(2019)Paszke, Gross, Massa, Lerer, Bradbury, Chanan, Killeen, Lin, Gimelshein, Antiga, et~al.]{paszke2019pytorch}
Adam Paszke, Sam Gross, Francisco Massa, Adam Lerer, James Bradbury, Gregory Chanan, Trevor Killeen, Zeming Lin, Natalia Gimelshein, Luca Antiga, et~al.
\newblock Pytorch: An imperative style, high-performance deep learning library.
\newblock \emph{Advances in neural information processing systems}, 32, 2019.

\bibitem[Preuer et~al.(2018)Preuer, Renz, Unterthiner, Hochreiter, and Klambauer]{preuer2018frechet}
Kristina Preuer, Philipp Renz, Thomas Unterthiner, Sepp Hochreiter, and Gunter Klambauer.
\newblock Fr{\'e}chet chemnet distance: a metric for generative models for molecules in drug discovery.
\newblock \emph{Journal of chemical information and modeling}, 58\penalty0 (9):\penalty0 1736--1741, 2018.

\bibitem[Ramakrishnan et~al.(2014)Ramakrishnan, Dral, Rupp, and Von~Lilienfeld]{ramakrishnan2014quantum}
Raghunathan Ramakrishnan, Pavlo~O Dral, Matthias Rupp, and O~Anatole Von~Lilienfeld.
\newblock Quantum chemistry structures and properties of 134 kilo molecules.
\newblock \emph{Scientific data}, 1\penalty0 (1):\penalty0 1--7, 2014.

\bibitem[Sastry et~al.(2011)Sastry, Dixon, and Sherman]{122}
G~Madhavi Sastry, Steven~L Dixon, and Woody Sherman.
\newblock Rapid shape-based ligand alignment and virtual screening method based on atom/feature-pair similarities and volume overlap scoring.
\newblock \emph{Journal of chemical information and modeling}, 51\penalty0 (10):\penalty0 2455--2466, 2011.

\bibitem[Schneider \& Fechner(2005)Schneider and Fechner]{111&121}
Gisbert Schneider and Uli Fechner.
\newblock Computer-based de novo design of drug-like molecules.
\newblock \emph{Nature Reviews Drug Discovery}, 4\penalty0 (8):\penalty0 649--663, 2005.

\bibitem[Shi et~al.(2020)Shi, Xu, Zhu, Zhang, Zhang, and Tang]{shi2020graphaf}
Chence Shi, Minkai Xu, Zhaocheng Zhu, Weinan Zhang, Ming Zhang, and Jian Tang.
\newblock Graphaf: a flow-based autoregressive model for molecular graph generation.
\newblock \emph{arXiv preprint arXiv:2001.09382}, 2020.

\bibitem[Simonovsky \& Komodakis(2018)Simonovsky and Komodakis]{112}
Martin Simonovsky and Nikos Komodakis.
\newblock Graphvae: Towards generation of small graphs using variational autoencoders.
\newblock In \emph{Artificial Neural Networks and Machine Learning--ICANN 2018: 27th International Conference on Artificial Neural Networks, Rhodes, Greece, October 4-7, 2018, Proceedings, Part I 27}, pp.\  412--422. Springer, 2018.

\bibitem[Taylor et~al.(2022)Taylor, Kardas, Cucurull, Scialom, Hartshorn, Saravia, Poulton, Kerkez, and Stojnic]{taylor2022galactica}
Ross Taylor, Marcin Kardas, Guillem Cucurull, Thomas Scialom, Anthony Hartshorn, Elvis Saravia, Andrew Poulton, Viktor Kerkez, and Robert Stojnic.
\newblock Galactica: A large language model for science.
\newblock \emph{arXiv preprint arXiv:2211.09085}, 2022.

\bibitem[Vaswani(2017)]{vaswani2017attention}
A~Vaswani.
\newblock Attention is all you need.
\newblock \emph{Advances in Neural Information Processing Systems}, 2017.

\bibitem[Vignac et~al.(2022)Vignac, Krawczuk, Siraudin, Wang, Cevher, and Frossard]{vignac2022digress}
Clement Vignac, Igor Krawczuk, Antoine Siraudin, Bohan Wang, Volkan Cevher, and Pascal Frossard.
\newblock Digress: Discrete denoising diffusion for graph generation.
\newblock \emph{arXiv preprint arXiv:2209.14734}, 2022.

\bibitem[Wang et~al.(2024)Wang, Wang, Wang, Cao, A~Saurous, and Kim]{wang2024grammar}
Bailin Wang, Zi~Wang, Xuezhi Wang, Yuan Cao, Rif A~Saurous, and Yoon Kim.
\newblock Grammar prompting for domain-specific language generation with large language models.
\newblock \emph{Advances in Neural Information Processing Systems}, 36, 2024.

\bibitem[Yao et~al.(2024)Yao, Wang, Zhang, Qin, Zhang, Chu, Yang, Zhu, and Mei]{yao2024exploring}
Yang Yao, Xin Wang, Zeyang Zhang, Yijian Qin, Ziwei Zhang, Xu~Chu, Yuekui Yang, Wenwu Zhu, and Hong Mei.
\newblock Exploring the potential of large language models in graph generation.
\newblock \emph{arXiv preprint arXiv:2403.14358}, 2024.

\bibitem[You et~al.(2018)You, Ying, Ren, Hamilton, and Leskovec]{you2018graphrnn}
Jiaxuan You, Rex Ying, Xiang Ren, William Hamilton, and Jure Leskovec.
\newblock Graphrnn: Generating realistic graphs with deep auto-regressive models.
\newblock In \emph{International conference on machine learning}, pp.\  5708--5717. PMLR, 2018.

\bibitem[Zang \& Wang(2020)Zang and Wang]{zang2020moflow}
Chengxi Zang and Fei Wang.
\newblock Moflow: an invertible flow model for generating molecular graphs.
\newblock In \emph{Proceedings of the 26th ACM SIGKDD international conference on knowledge discovery \& data mining}, pp.\  617--626, 2020.

\end{thebibliography}
\bibliographystyle{iclr2025_conference}

\clearpage

\appendix

\section{Additional experiments results}\label{app:res}

Here are additional experiment results on QM9 and ZINC250k datasets. The \textbf{Neighborhood Subgraph Pairwise Distance Kernel (NSPDK) Maximum Mean Discrepancy (MMD)}~\citep{costa2010fast} evaluates the difference between generated and test molecules, accounting for both node and edge features. 
\textbf{Uniqueness} refers to the percentage of valid molecules that are distinct from each other. \textbf{Validity}, \textbf{FCD}, \textbf{Novelty}, and \textbf{Scaf} have been introduced before.

\begin{table}[ht]
    \centering
    \caption{Generation results on the QM9 dataset. We report the mean of 3 different
runs. The best results are highlighted in bold. The second-best results are highlighted in underline.}
    \label{tab:qm9}
    \begin{tabular}{l|cccccc}
    \toprule
    \textbf{Methods}  & \textbf{Valid} (\%)$\uparrow$ & \textbf{FCD} $\downarrow$ & \textbf{NSPDK} $\downarrow$ & \textbf{Scaf} $\uparrow$ & \textbf{Unique} (\%)$\uparrow$ & \textbf{Novelty} (\%)$\uparrow$ \\ \midrule

    MoFlow            & 91.36 & 4.467 & 0.017 & 0.1447 & \underline{98.65} & \underline{94.72} \\
    GraphAF           & 74.43 & 5.625 & 0.021 & 0.3046 & 88.64 & 86.59 \\
    GraphDF           & 93.88 & 10.928 & 0.064 & 0.0978 & 98.58 & \textbf{98.54} \\ \midrule
    EDP-GNN           & 47.52 & 2.680 & 0.005 & 0.3270 & \textbf{99.25} & 86.58 \\
    GDSS              & 95.72 & 2.900 & 0.003 & 0.6983 & 98.46 & 86.27 \\
    DiGress           & 98.19 & \textbf{0.095} & \underline{0.0003} & \underline{0.9353} & 96.67 & 25.58 \\
    Grum              & \textbf{99.69} & \underline{0.108} & \textbf{0.0002} & \textbf{0.9449} & 96.90 & 24.15 \\ \midrule
    Ours              & \underline{99.47} & 0.815 & 0.002 & 0.9112 & 89.57 & 88.29 \\ \bottomrule
    \end{tabular}
\end{table}

\begin{table}[ht]
    \centering
    \caption{Generation results on the ZINC250k dataset. We report the mean of 3 different runs. The best results are highlighted in bold. The second-best results are highlighted in underline.}
    \label{tab:zinc}
    \begin{tabular}{l|cccccc}
    \toprule
    \textbf{Methods}  & \textbf{Valid} (\%)$\uparrow$ & \textbf{FCD} $\downarrow$ & \textbf{NSPDK} $\downarrow$ & \textbf{Scaf} $\uparrow$ & \textbf{Unique} (\%)$\uparrow$ & \textbf{Novelty} (\%)$\uparrow$ \\ \midrule

    MoFlow            & 63.11 & 20.931 & 0.046 & 0.0133 & \textbf{99.99} & \textbf{100.00} \\
    GraphAF           & 68.47 & 16.023 & 0.044 & 0.0672 & 98.64 & 99.99 \\
    GraphDF           & 90.61 & 33.546 & 0.177 & 0.0000 & 99.63 & \textbf{100.00} \\ \midrule
    EDP-GNN           & 82.97 & 16.737 & 0.049 & 0.0000 & 99.79 & \textbf{100.00} \\
    GDSS              & 97.01 & 14.656 & 0.019 & 0.0467 & 99.64 & \textbf{100.00} \\
    DiGress           & 94.99 & 3.482 & \underline{0.0021} & 0.4163 & \underline{99.97} & 99.99 \\
    Grum              & \textbf{98.65} & \textbf{2.257} & \textbf{0.0015} & \underline{0.5299} & \underline{99.97} & 99.98 \\ \midrule
    Ours    & \underline{98.03} & \underline{2.445} & 0.0049 & \textbf{0.6062} & 94.69 & \textbf{100.00} \\ \bottomrule
    \end{tabular}
\end{table}


\end{document}